\documentclass{article}
\usepackage{iclr2027_conference,times}

\usepackage{amsmath,amsfonts,bm}

\def\eqref#1{equation~\ref{#1}}

\def\1{\bm{1}}

\DeclareMathAlphabet{\mathsfit}{\encodingdefault}{\sfdefault}{m}{sl}
\SetMathAlphabet{\mathsfit}{bold}{\encodingdefault}{\sfdefault}{bx}{n}

\usepackage{booktabs}
\usepackage{microtype}
\usepackage[table]{xcolor}
\usepackage{amssymb}
\usepackage{float}
\usepackage{multirow, multicol}
\usepackage{graphicx}
\usepackage{adjustbox}
\usepackage{hyperref}
\usepackage{tikz}
\usetikzlibrary{arrows.meta,calc,positioning}

\usepackage{silence}
\title{Natural Gradient Learning in Predictive\\Coding through Lateral Circuits}

\author{Lhea Beumer$^{1,2,*}$, Sander de Haan$^{1,2,*}$, Niclas Pokel$^{1,2,*}$, \\ Pau Vilimelis Aceituno$^{1,2}$, Benjamin F. Grewe$^{1,2}$ \\
$^{1}$Institute of Neuroinformatics, UZH \& ETH Zurich,
$^{2}$ETH AI Center,
$^{*}$equal contribution
}

\iclrfinalcopy

\begin{document}
\maketitle

\begin{abstract}
Predictive coding (PC) trains neural networks through local energy-minimization dynamics without depending on a global error signal that backpropagation (BP) requires. Although PC is a promising alternative algorithm for efficient learning, the computational cost of inference dynamics currently outweighs the resulting learning benefits in practice. In this work, we take inspiration from optimization theory to extract more learning benefit from each inference phase. We introduce \textit{Natural PC}, which augments standard PC with a lateral circuit, enabling exact and efficient natural gradient learning locally in space and time. Our lateral circuit learns the activity correlations that determine the Fisher information matrix and implicitly computes its inverse, avoiding expensive matrix inversion. Across all tested image reconstruction benchmarks, Natural PC achieves state of the art in terms of sample and compute efficiency compared to existing PC approaches. Most notably, Natural PC scales to ImageNet and matches the performance of BP trained with AdamW and various second-order optimizers at an order of magnitude lower computational cost while also using two orders of magnitude fewer training samples. Our results demonstrate that local natural gradient learning makes PC competitive with BP in both sample efficiency and computational cost in reconstruction tasks.

\end{abstract}

\section{Introduction}

Predictive coding (PC) is a prominent theory of cortical computation, where higher cortical areas predict lower-level sensory inputs, unifying perception and learning as the local minimization of prediction errors \citep{rao1999predictive,friston2005theory,friston2008hierarchical,friston2010free,spratling2010predictive,bastos2012canonical,shipp2016neural}.
The local computation of prediction errors in PC provides a mechanism to solve the credit assignment problem in the brain \citep{rumelhart1986learning,lillicrap2020backpropagation} and has therefore motivated the development of algorithms for training artificial neural networks.
Although PC can approximate or exactly implement backpropagation (BP) under specific conditions \citep{whittington2017approximation,song2020can,salvatori2021predictivecodingexactbackpropagation,rosenbaum2022relationship,millidge2022backpropagation,innocenti2026infinitewidth,seely2026augmentedlagrangianpredictivecoding}, recent work suggests that PC can also provide computational advantages over BP, including fewer weight updates \citep{alonso2022theoretical,alonso2024understanding}, improved online and continual learning \citep{Songetal2024}, faster escape from saddle points \citep{innocenti2024strictsaddlesenergylandscape}, and greater sample efficiency \citep{oliviers2026understandingsampleefficiencypredictive, vectornetworks}. Together, these properties make PC a promising foundation for efficient learning in neural networks.

Despite this promise, standard PC remains difficult to apply in practice, since even relatively simple problems can require substantial computational resources \citep{pinchetti2025benchmarkingpredictivecodingnetworks}.
Unlike BP, which straightforwardly computes synaptic updates by differentiating a global loss function, PC treats the network as a dynamical system and requires an inference process to solve the dynamics before synaptic weights are updated \citep{Songetal2024,aceituno2025challenging}.
Once inference has minimized prediction errors and the dynamics have reached equilibrium, the synaptic weights are updated to reduce the residual prediction errors at that equilibrium \citep{whittington2017approximation}.
However, reaching equilibrium often requires hundreds or thousands of inference iterations per sample, and PC becomes practical only when the trade-off between inference costs and learning benefits is favorable.

Recent work has sought to improve this trade-off by reducing inference costs or increasing the learning benefit from inference.
To reduce inference costs, some approaches learn activity estimates from the input or initialize activities from progress on earlier samples to provide inference with a more informative starting point \citep{8444980,10.1371/journal.pcbi.1011280,pinchetti2026fasterpredictivecodingnetworks}.
Other approaches improve the learning benefit from inference by mitigating ill-conditioned inference \citep{NEURIPS2025_7ea5ad70}, balancing prediction errors across layers \citep{qi2025trainingdeeperpredictivecoding}, preventing prediction errors from vanishing or exploding \citep{ICLR2026_319e084c}, or improving credit assignment through local constraints \citep{seely2026augmentedlagrangianpredictivecoding}.
Although these existing approaches improve inference and learning, their synaptic updates do not follow the local geometry of the model distribution \citep{amari1987differential,6790500}, leaving open an information-geometric route to more efficient learning.

In this work, we introduce \textit{Natural PC}, an extension of standard PC with a lateral interneuron circuit enabling natural gradient learning.
We first derive an exact layerwise form of the natural gradient for PC, identifying the activity correlations required for geometry-aware synaptic updates.
We then show that lateral weights learn these activity correlations and that the dynamics of the interneuron circuit implicitly invert the correlations during inference, removing the need for explicit and expensive matrix inversion.
We theoretically show that the natural gradient of the PC learning objective coincides with Newton and Gauss--Newton updates, explaining how Natural PC can obtain more learning benefit from each costly inference phase.
As the lateral circuit controls both the geometry and learning rate of the network, we remove both the learning rate hyperparameter and the weight optimizer.
Across multiple image reconstruction datasets, Natural PC achieves state of the art in terms of reconstruction error, sample efficiency, and computational cost compared to existing PC methods.
Most notably, Natural PC scales to ImageNet-1k and matches the performance of BP trained with AdamW and various second-order optimizers at an order of magnitude lower computational cost while also using two orders of magnitude fewer training samples.
 \section{Natural Predictive Coding}
\label{sec:natural-pc}

We now introduce \emph{Natural PC}, which extends PC with a lateral interneuron circuit that implements natural gradient learning.
We first provide a background on standard PC (Sec.~\ref{sec:pc-prediction-errors}), and derive the natural gradient of the PC learning objective (Sec.~\ref{sec:pc-natural-gradient}).
We then introduce a lateral circuit that allows the network to perform natural gradient descent (Sec.~\ref{sec:lateral-natural-gradient}), and show that natural gradient learning in PC coincides with Newton and Gauss--Newton optimization (Sec.~\ref{sec:gn-regularization}).

\subsection{Predictive Coding Background}
\label{sec:pc-prediction-errors}

We formalize the PC network as a directed Gaussian generative model of depth $L$, following the standard formulation \citep{rao1999predictive,friston2005theory,bogacz2017tutorial,buckley2017free}.
Let $z_l$ denote the activity at layer $l$, with $z_0$ denoting the observed input; let $\phi$ denote a smooth element-wise nonlinearity; let $W_l$ contain the top-down weights from layer $l$ to layer $l-1$; and let $\Pi_l$ denote the precision, or inverse residual covariance, of the corresponding conditional distribution.
The PC network (see Fig.~\ref{fig:sketch-pc-overview}a) factorizes as,

\begin{equation}
p(z_{0:L}\mid \theta)
= p(z_L) \prod_{l=1}^{L} p(z_{l-1} \mid z_l; W_l, \Pi_l),
\ \
p(z_{l-1} \mid z_l; W_l, \Pi_l)
= \mathcal{N}\!\bigl(z_{l-1}; W_l\phi(z_l), \Pi_l^{-1}\bigr),
\label{eq:genmodel}
\end{equation}

with learnable weights $\theta=\{W_l\}_{l=1}^{L}$.
Intuitively, the precision $\Pi_l$ specifies the weight assigned to each prediction-error component, leading to larger penalties along low-variance dimensions than along noisy dimensions.
For simplicity and to isolate the learning geometry, we hold the precision parameters fixed, instead of inferring or learning these parameters \citep{friston2008hierarchical,ofner2021predictivecodingprecisionnatural}.

PC learns by alternating the inference of the activities while keeping the weights fixed, and the learning of the weights once the activities have settled.
For a clamped input $z_0$, inference minimizes the precision-weighted energy $\mathcal E$ by gradient descent on the activities \citep{friston2005theory},

\begin{equation}
z_{1:L}^{*}=\arg\min_{z_{1:L}}\mathcal{E}(z_{0:L},\theta),
\qquad
\mathcal{E}(z_{0:L},\theta)=\tfrac{1}{2}\sum_{l=1}^{L}\varepsilon_l^{\!\top}\Pi_l\varepsilon_l,
\qquad
\varepsilon_l=z_{l-1}-W_l\phi(z_l).
\label{eq:freeenergy}
\end{equation}

where $\varepsilon_l$ is the top-down prediction error of $z_{l-1}$.
Once inference has reached equilibrium $(\varepsilon^*,z^*)$, the weights are updated following the local Hebbian rule \citep{whittington2017approximation,millidge2021predictive},

\begin{equation}
\Delta W_l^{\mathrm{PC}}
=-\eta\left.\nabla_{W_l}\mathcal{E}\right|_{z^*}
=\boxed{\eta\,\Pi_l\varepsilon_l^*\phi(z_l^*)^{\top}}.
\label{eq:pc-update}
\end{equation}

Minimizing the energy with respect to the weights of each layer with the equilibrium activities held fixed is equivalent to maximizing $\log p(z_{l-1}^*\mid z_l^*;W_l,\Pi_l)$ under the Gaussian model in Eq.~\ref{eq:genmodel}.
Intuitively, each layer treats the equilibrium activity $\phi(z_l^{*})$ as a fixed input and the equilibrium activity $z_{l-1}^{*}$ of the layer below as a noisy target, such that learning in each layer is a linear regression of the layer below onto the layer above, with the prediction error as the residual.

\subsection{Natural Gradient Learning for Predictive Coding}
\label{sec:pc-natural-gradient}

We now identify a limitation of the standard PC learning rule that restricts the effectiveness of learning after inference.
The learning rule in Eq.~\ref{eq:pc-update} follows the Euclidean gradient and does not account for differences in curvature across parameters.
Because the objective can be much more curved in some directions than in others, a descent step must be small enough to avoid overshooting along the most strongly curved directions, leading to slow progress along flatter ones \citep{6790500}.
The single Euclidean weight update performed after inference generally does not fully maximize the likelihood given the equilibrium activities, and completing the maximization would generally require further weight updates, thereby limiting the learning benefit obtained from each costly inference phase.
We observe that the natural metric of the model would allow a single appropriately scaled natural gradient step to reproduce the exact maximization \citep{amari1995information}, eliminating the need for repeated weight updates (see Fig.~\ref{fig:sketch-pc-overview}b).

To derive the natural gradient for PC, we rescale the energy gradient by the inverse Fisher information matrix \citep{amari1987differential,6790500},

\begin{equation}
\widetilde{\nabla}_{\theta}\mathcal{E}
=\mathcal{F}^*(\theta)^{-1}\nabla_{\theta}\mathcal{E},
\qquad
\mathcal{F}^*(\theta)
=\mathbb{E}\!\left[
\nabla_{\theta}\log p\,\nabla_{\theta}\log p^{\top}
\right],
\label{eq:natural-gradient-definition}
\end{equation}

where $\mathcal{F}^*(\theta)$ is the Fisher of the layerwise Gaussian regressions.
Intuitively, the Fisher measures how strongly parameter changes alter the model distribution in each direction, and the inverse Fisher therefore rescales update magnitudes to give smaller steps along sensitive directions and larger steps along insensitive directions, addressing the curvature imbalance that limits learning in the Euclidean metric.

To compute the Fisher, we hold $z_l^*$ fixed, average over $z_{l-1}\sim p(z_{l-1}\mid z_l^*;W_l,\Pi_l)$, and then average over observed inputs.
We obtain (see App.~\ref{app:ng-derivation}),

\begin{equation}
\mathcal{F}_l^{*}=G_l^*\otimes\Pi_l,
\qquad
G_l^*\triangleq\mathbb{E}_{z_0\sim p_{\mathrm{data}}}
\!\left[\phi(z_l^*)\phi(z_l^*)^{\top}\right],
\label{eq:fisher}
\end{equation}

where $\otimes$ denotes the Kronecker product and $G_l^*$ represents the second moment of the equilibrium presynaptic activities, which we refer to as the Gram matrix.
The learning objective thus factorizes across layers, making the Fisher block diagonal with one matrix $\mathcal F_l^*$ for each layer.
Substituting into the learning rule of Eq.~\ref{eq:pc-update} gives,

\begin{equation}
\Delta W_l^{\mathrm{NG}}
=-\eta\,\Pi_l^{-1}
\left(-\Pi_l\varepsilon_l^*\phi(z_l^*)^{\top}\right)G_l^{*-1}
=\boxed{\eta\,\varepsilon_l^*\phi(z_l^*)^{\top}G_l^{*-1}},
\label{eq:nat-grad-target}
\end{equation}

where the precision factor from the learning rule cancels with the corresponding factor of the inverse Fisher.
The natural geometry rescales the standard PC update according to correlations among presynaptic activities, allowing larger steps along directions where Euclidean learning progresses slowly without overshooting along more strongly curved directions. Intuitively, accounting for presynaptic correlations prevents correlated inputs from performing redundant weight updates, allowing larger steps along directions that would otherwise learn slowly.
The resulting weight update can therefore make more progress toward the likelihood maximum after each inference phase without requiring additional inference iterations.
We now show how a lateral circuit learns the Gram matrix and applies the inverse through local dynamics.

\subsection{Lateral Circuits Implement Natural Gradient Learning}
\label{sec:lateral-natural-gradient}

\begin{figure}[t]
    \centering
    \includegraphics[width=\textwidth]{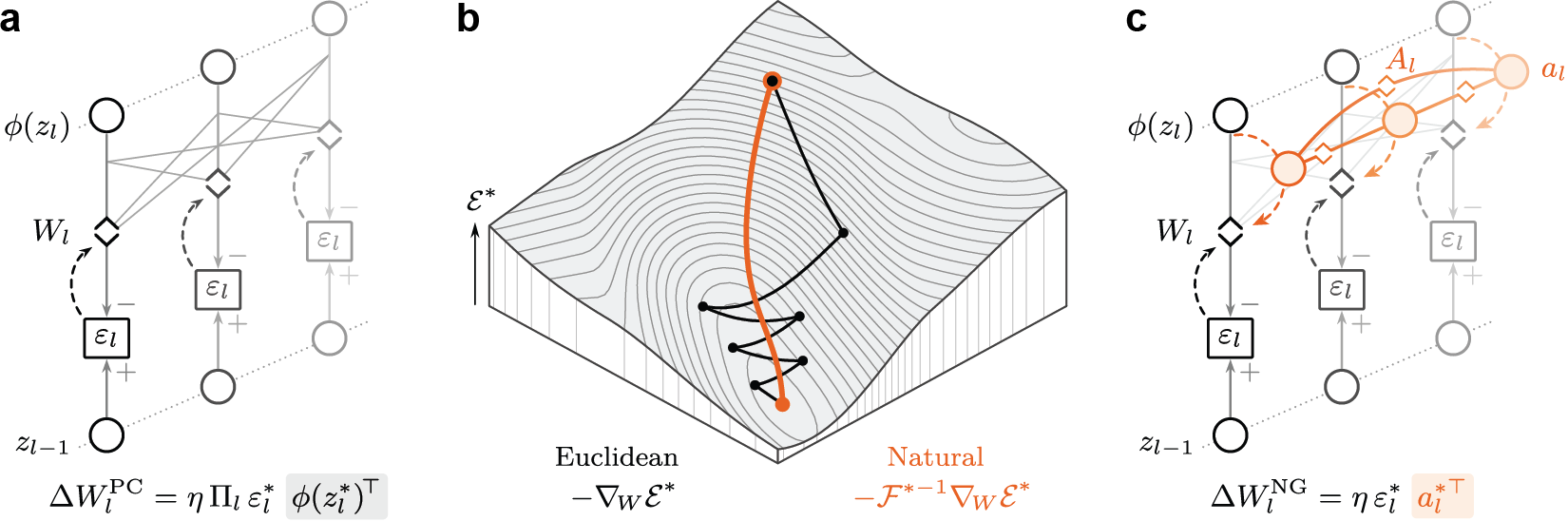}
    \caption{Natural Predictive Coding.
      \textbf{(a)} Standard PC. Layer $l$ sends $\phi(z_l)$ through $W_l$ (diamonds) to predict the layer below, and error neurons $\varepsilon_l$ compare the prediction with the activity. Dashed arrows carry plasticity signals.
      \textbf{(b)} Energy $\mathcal{E}$ over a slice of weight space at equilibrium activity. Euclidean descent (black) takes many steps across the valley, while the natural gradient (orange) reaches the minimum in one step.
      \textbf{(c)} Natural PC adds interneurons with activities $a_l$ that receive $\phi(z_l)$ and interact through lateral weights $A_l$ to obtain the natural geometry.}
  \label{fig:sketch-pc-overview}
\end{figure}

In order to implement natural gradient learning, we need to estimate the presynaptic correlations across training samples to construct the requisite Gram matrix, and apply its inverse to the presynaptic activities.
We introduce a lateral circuit consisting of a population of interneurons with activities $a_l$ and lateral weights $A_l$ for every layer, governed by,

\begin{equation}
\tau_a\dot a_l
=\phi(z_l)-A_l a_l,
\qquad
\tau_A\dot A_l
=\phi(z_l^*)\phi(z_l^*)^{\top}-A_l,
\label{eq:lateral-circuit}
\end{equation}

where the interneurons receive the activities from the hidden neurons, and the lateral weights connect the interneurons within each layer (Fig.~\ref{fig:sketch-pc-overview}c).
The interneuron dynamics evolve inside the inference phase, while the lateral weights update at equilibrium on the slower timescale of synaptic learning together with the feedforward weights.

The lateral weights accumulate presynaptic correlations across samples over time \citep{foldiak1990forming,pehlevan2015optimization,PehlevanHuChklovskii2015}, and thereby estimate the Gram matrix as an exponential moving average.
The interneuron dynamics apply the inverse of the learned matrix through a dynamical inversion \citep{NEURIPS2020_72619259}.
For stationary activity statistics and symmetric positive-definite $A_l$, the lateral dynamics and the interneuron equilibrium satisfy (see App.~\ref{app:lateral-circuit}),

\begin{equation}
A_l\longrightarrow G_l^*,
\qquad
a_l^*=A_l^{-1}\phi(z_l^*).
\label{eq:equilibrium-gram}
\end{equation}

The local synaptic rule of Natural PC uses the interneuron activities as the presynaptic factor when inference reaches the new equilibrium $(\varepsilon^*,z^*,a^*)$,

\begin{equation}
\Delta W_l^{\mathrm{NPC}}
=\eta\,\varepsilon_l^*a_l^{*\top}
=\boxed{\eta\,\varepsilon_l^*\phi(z_l^*)^{\top}A_l^{-1}}.
\label{eq:natural-pc-update}
\end{equation}

We thereby recover the natural gradient learning rule of Eq.~\ref{eq:nat-grad-target} exactly when $A_l=G_l^*$.
The lateral circuit thus implements natural gradient learning locally in space and time, where the approximation during training is determined by how closely $A_l$ tracks $G_l^*$, as we empirically verify in Sec.~\ref{sec:empirical_validation}.

\subsection{Natural PC Implements Newton Optimization}
\label{sec:gn-regularization}

We now ask which optimization principle Natural PC implements.
Unlike first-order gradient descent using gradients computed by BP, curvature-based second-order optimization accounts for how quickly the gradient changes along different weight directions, allowing larger steps toward a minimum along directions where gradient descent progresses slowly.
To recap, rescaling the gradient by the inverse Hessian, or by the inverse of a curvature approximation obtained by linearizing the predictions, is called Newton or Gauss--Newton optimization, respectively \citep{nocedal2006numerical}.

As we have seen, weight learning in PC minimizes the Gaussian negative log-likelihood, where the learning objective and its curvature described by the Hessian $\mathcal H$ are,

\begin{equation}
\bar{\mathcal E}_l(W_l)
=-\mathbb E
\big[\log p(z_{l-1}^*\mid z_l^*;W_l,\Pi_l)\big],
\qquad
\mathcal H_l^*
=\nabla^2\bar{\mathcal E}_l
=\mathbb E
\big[J_l^{*\top}\Pi_lJ_l^*\big],
\label{eq:fixed-activity-objective}
\end{equation}

where $J_l^*=\phi(z_l^*)^\top\otimes I$ is the Jacobian of the layer predictions with respect to the weights.
Since the predictions are linear in the weights, differentiating the objective twice gives exactly the Jacobian product defining the Gauss--Newton matrix $\mathcal G$.
Additionally, for the Gaussian likelihood with fixed precision, the Fisher equals the same Jacobian product \citep{martens-2020}.
Consequently, we obtain the following unification (see App.~\ref{app:curvature-identities}),

\begin{equation}
\mathcal H_l^*
=\mathcal F_l^*
=\mathcal G_l^*
=G_l^*\otimes\Pi_l.
\label{eq:exact-layerwise-curvature}
\end{equation}

Natural PC therefore implements Newton and Gauss--Newton updates in addition to natural gradient learning.

Further, since the learning objective of each layer is quadratic in the weights, our natural gradient and Newton update reaches the likelihood maximum with a unit step size (App.~\ref{app:exact-maximization}), allowing us to remove the learning rate hyperparameter.
As learning progresses and $A_l\to G_l^*$, one update of unit size lands on the least-squares solution \citep{plackett1950some,haykin2002adaptive}, whereas Euclidean learning would approach this solution over many small steps (Fig.~\ref{fig:sketch-pc-overview}b).
Setting $A_l=I$ recovers standard PC, which takes these Euclidean steps and thereby requires a learning rate small enough to avoid overshooting along the most strongly curved direction.
The lateral weights thus determine both the geometry and the learning rate of the synaptic update. \section{Empirical Validation of the Learning Theory}
\label{sec:empirical_validation}

We now empirically show how the lateral circuit of Natural PC aligns with the predictions of the learning theory by answering three questions.
Do the lateral weights recover the natural gradient during training (Sec.~\ref{sec:val-gram})?
Do the lateral weights align with the full data distribution required for the natural gradient (Sec.~\ref{sec:val-data})?
Does the lateral circuit support unit natural gradient steps (Sec.~\ref{sec:val-unit})?

To answer these questions, we train a three-layer generative model on the 50,000 training images of CIFAR-10, with 128 hidden neurons per layer, $\phi=\tanh$, isotropic precisions $\Pi_l=\sigma^{-2}I$ with $\sigma^2=0.1$, and a unit Gaussian prior on the top layer.
Mini-batches contain $B=200$ images, giving 250 weight updates per epoch.
The lateral weights are initialized at $A_l=I$ and integrate the activities with $\tau_A=20$ (see Sec.~\ref{sec:val-data}).
Further experimental details are provided in App.~\ref{app:adaptive-rate}.

\subsection{The Lateral Weights Recover Natural Gradient Directions}
\label{sec:val-gram}

\begin{figure}[t]
  \centering
  \includegraphics[width=\linewidth]{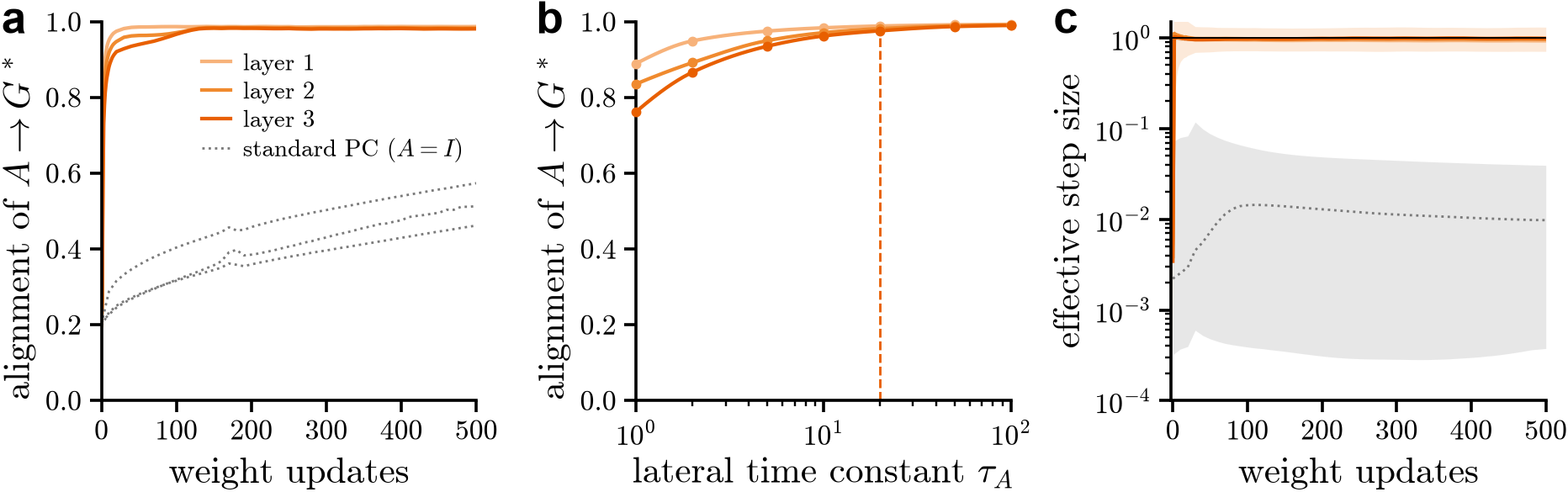}
  \caption{The lateral circuit learns the geometry of natural gradient updates.
  A three-layer generative model is trained on CIFAR-10.
  \textbf{(a)} Alignment of the lateral weights $A_l$ with the Gram matrix $G_l^*$ over training.
  \textbf{(b)} Alignment after training as a function of the lateral time constant $\tau_A$. The dashed line marks $\tau_A=20$.
  \textbf{(c)} Effective step size over training. Median and 5th to 95th percentile across directions and layers. The black line indicates the unit step.
  }
  \label{fig:learning-theory}
\end{figure}

We first test whether the learned lateral weights produce the directions required by the natural gradient.
We measure alignment using an activity-weighted cosine similarity between the directions produced by $A_l^{-1}$ and $G_l^{*-1}$,

\begin{equation}
\mathrm{align}(A_l,G_l^*)
=\frac{\langle A_l^{-1},G_l^{*-1}\rangle}
{\|A_l^{-1}\|\,\|G_l^{*-1}\|},
\label{eq:alignment-cosine}
\end{equation}

where the inner product and norms are weighted by $G_l^*$, measuring directional agreement independently of step size.
We observe that alignment starts near 0.2 at random network initialization, rises to 0.88--0.95 within 10 weight updates, and becomes nearly perfect across layers within 100 updates, reaching 0.97--0.99 and increasing to 0.98--0.99 after 500 updates (Figure~\ref{fig:learning-theory}a).
Standard PC without the lateral circuit never reaches comparable alignment and remains below 0.6 averaged across layers.
The lateral weights therefore recover the natural gradient directions almost perfectly during training.

\subsection{The Lateral Weights Align with the Full Data Distribution}
\label{sec:val-data}

The natural gradient depends on activity correlations across the full data distribution $p_{\mathrm{data}}$, as shown in Eq.~\ref{eq:fisher}.
However, each lateral weight update observes only a single batch, so the time constant $\tau_A$ controls an exponential moving average across batches to estimate the full-data activity statistics.
We therefore vary $\tau_A$ and measure the alignment of $A_l$ and $G_l^*$ after training.
We observe that at $\tau_A=1$, where the lateral weights track only the current batch, alignment reaches approximately 0.76--0.89 across layers.
Increasing to $\tau_A=20$ raises alignment to approximately 0.98--0.99, with similar values up to $\tau_A=100$ (Fig.~\ref{fig:learning-theory}b).
Smaller time constants adapt more quickly as the feedforward weights learn, whereas larger time constants average over more batches reduce fluctuations in the estimated activity correlations.
The lateral weights can thus recover the activity statistics of the full data distribution required for the natural gradient by integrating local correlations across batches.

\subsection{The Lateral Circuit Enables Unit Learning Steps}
\label{sec:val-unit}

We finally test whether the lateral circuit approaches unit natural gradient steps during training.
We compute the effective step size for every direction $i$,

\begin{equation}
s_{l,i}=\eta_l\lambda_i\!\left(A_l^{-1}G_l^*\right),
\label{eq:effective-step}
\end{equation}

where $\eta_l=1$ for Natural PC and $s_{l,i}=1$ denotes an exact unit step.
We plot the median and the 5th to 95th percentile across directions of the effective step size for Natural PC and standard PC in Figure~\ref{fig:learning-theory}c (Appendix~\ref{app:effective-step}).
Starting near 0.004, the layer medians of Natural PC reach 0.9 to 1.1 within two updates and the spread across directions narrows to a factor of two within 100 updates, while standard PC retains a spread of two orders of magnitude, indicative of the ill-conditioning of the Euclidean metric.
The lateral circuit therefore approaches unit learning steps in natural gradient learning. \section{Benchmarks}
\label{sec:bench}
We now test whether the natural gradient of the lateral circuit translates into faster and better learning. We first show that Natural PC substantially outperforms all tested variants of PC, as well as all tested BP methods trained with different optimizers, in terms of sample and compute efficiency on Imagenet-1k reconstruction (Sec.~\ref{sec:bench-imagenet}). We then show that the performance and efficiency gains of Natural PC also generalize to all other tested smaller image reconstruction benchmarks (Sec.~\ref{sec:multiple_datasets}).

\subsection{Natural PC Achieves Fast and Sample Efficient Learning}
\label{sec:bench-imagenet}

We test Natural PC on ImageNet-1k \citep{deng2009imagenet} at $224\times224$ resolution with a convolutional generative model. We implement a top-down VGG-style decoder \citep{simonyan2015very} that maps the images through nine $3\times3$ convolutional layers with an $8{,}192$-dimensional bottleneck using $\phi=\tanh$, comprising $212.6$M parameters. To the best of our knowledge, this network is roughly four times larger than the current largest PC model ever successfully trained \citep{kerjan2026trainingpredictivecodingnetwork}. The image is clamped at the bottom and all hidden layers are inferred with $30$ Nesterov-accelerated energy settling steps \citep{nesterov1983solving, pinchetti2025benchmarkingpredictivecodingnetworks, rao2026encyclopedia}. The lateral circuit solves its linear system once per batch rather than settling over many iterations, and the lateral weights follow a running average of the co-activity of the $3\times3$ input patches of each layer.

We compare against existing PC models including standard PC \citep{whittington2017approximation}, incremental PC (iPC) \citep{salvatori2024stablefastfullyautomatic}, error-based PC (ePC) \citep{goemaere2026epcfastdeeppredictive}, bidirectional PC (bPC) \citep{oliviers2025bidirectionalpredictivecoding}, and augmented Lagrangian PC (PC-ALM) \citep{seely2026augmentedlagrangianpredictivecoding}. We additionally compare against a number of BP methods trained with different weight optimizers at tuned learning rates, including SGD and AdamW \citep{kingma2015adam,loshchilov2019decoupled}, as well as second-order optimizers comparable to our method, including Muon \citep{jordan2024muon} and SOAP \citep{vyas2025soap}. For the BP baselines, we train an autoencoder whose decoder is the same generative model and whose encoder is the same architecture flipped. We report the reconstruction error (MSE) on $5{,}000$ class-balanced validation images decoded from the top-layer activity, which is inferred with $300$ Nesterov steps for the PC models and given by the decoder for the BP models. Reported training FLOPs include the inference steps of the PC models and the computations of the lateral circuit in Natural PC. We give detailed description of the training procedure of each model and  provide additional details of all experimental settings in  App.~\ref{app:imagenet-protocol}.

\begin{figure}[t]
  \centering
  \includegraphics[width=\linewidth]{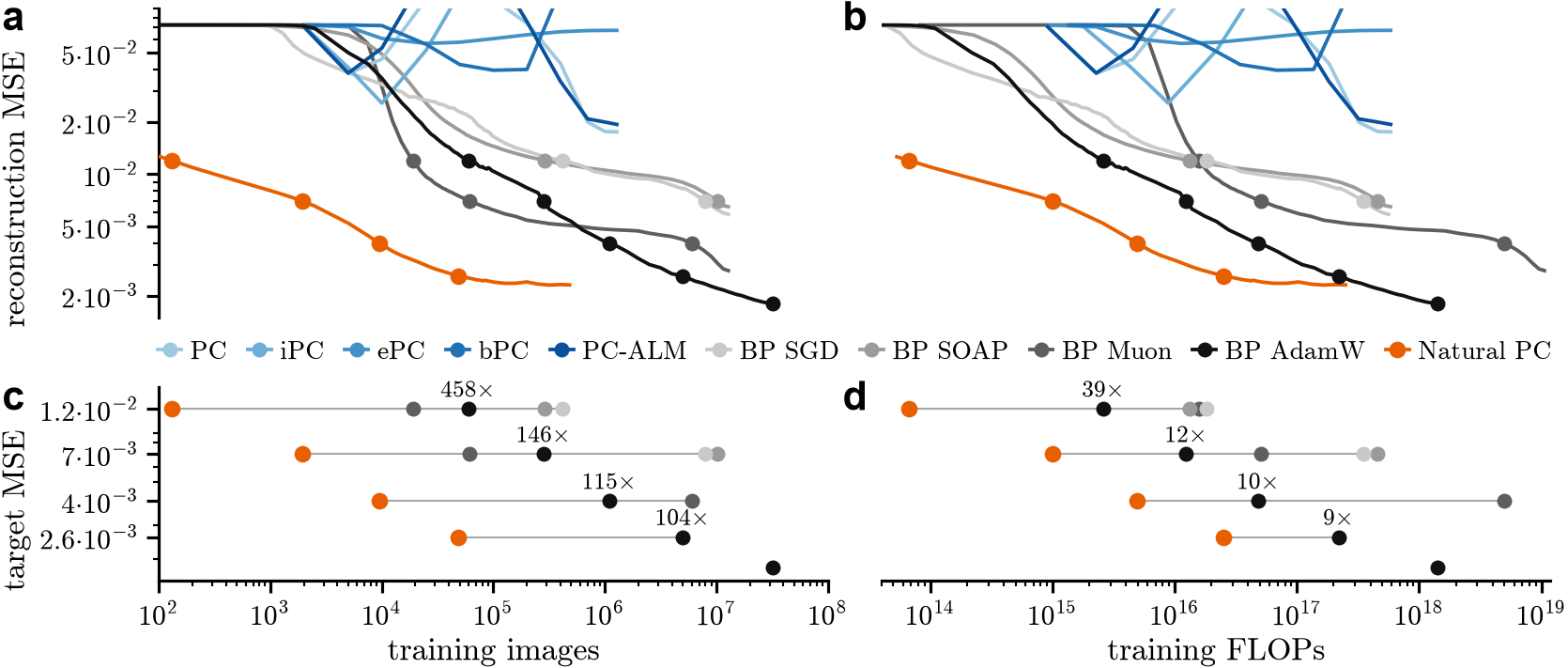}
  \caption{The sample and compute efficiency of Natural PC on ImageNet reconstruction versus PC and BP baselines.
  \textbf{(a-b)} Held-out reconstruction MSE against training images and FLOPs. Dots mark where each curve first reaches a target MSE of (c).
  \textbf{(c-d)} Training images and FLOPs each method needs to reach a target MSE. Multipliers give the ratio of BP AdamW to Natural PC. The bottom row marks the final MSE of BP AdamW.}
  \label{fig:imagenet}
\end{figure}

We summarize our results in Fig.~\ref{fig:imagenet}. Natural PC reaches a reconstruction MSE of $2.3\cdot10^{-3}$ after $5\cdot10^{5}$ training images, less than half an epoch. iPC and bPC diverge at this scale, and PC, ePC and PC-ALM stall above the highest target MSE of $1.2\cdot10^{-2}$ within one epoch. Thus, none of the PC baselines are able to scale to this task. Every BP baseline needs orders of magnitude more training images than Natural PC to reach each target MSE, and BP with AdamW needs $460$ to $100$ times more. BP with AdamW matches the final MSE of Natural PC only after $8.3\cdot10^{6}$ images, $17$ times more, and reaches a lower MSE of $1.8\cdot10^{-3}$ after $25$ epochs. Muon, SOAP, and SGD do not reach the final MSE of Natural PC. Contrary to our expectations, neither second-order optimizer competes with AdamW, as Muon ends at $2.8\cdot10^{-3}$ and SOAP at $6.5\cdot10^{-3}$, and both spend more training FLOPs than AdamW to reach every target MSE.

Strikingly, Natural PC reaches every target with $39$ to $9$ times fewer training FLOPs than BP with AdamW, even though Natural PC spends $12$ times more FLOPs per batch, which includes the $30$ settling steps per batch. Moreover, Natural PC sets its own step size from the local curvature and therefore needs no learning rate, whereas every other baseline required a sweep over learning rates before reaching these results, further widening the gap in compute cost. However, Natural PC stops improving after about $10^{5}$ images and slowly degrades beyond $5\cdot10^{5}$ images, whereas BP with AdamW keeps improving. We attribute Natural PC's early plateau to each hidden layer settling to absorb the error left unexplained by the layer above. Consequently, reconstruction from the equilibrium activity continues to improve without corresponding improvements in reconstruction decoded from the top layer. This is a known open problem in hierarchical latent-variable models, in which the lowest layers alone can explain the data at the optimum \citep{zhao2017learning}. In addition, each layer learns only from equilibrium activities but is evaluated during decoding from top-down predictions, creating a mismatch between the inputs used for learning versus those used for reconstruction.

\subsection{Natural PC on Standard PC Benchmarks}\label{sec:multiple_datasets}

For the smaller datasets, we follow \citet{pinchetti2025benchmarkingpredictivecodingnetworks} for benchmarking PC networks. We compare Natural PC against the same PC models as tested on ImageNet. All PC methods train the generative model of Eq.~\ref{eq:genmodel} with three hidden layers of $128$ units, and $\phi=\tanh$. We compare five datasets comprising MNIST, FMNIST, CIFAR-10, STL-10, and Tiny ImageNet, using mini-batches of $200$ samples. We report the MSE on $512$ held-out samples, along with the number of epochs and weight updates required for each method to plateau, defined by reaching $2\%$ of the final error. We train each method to its own plateau because the methods converge to different errors, and thus a shared target error would not be reached by all methods. Additionally, a shared weight update budget would compare methods at different stages of training and, therefore, would not allow for a fair comparison of the MSE. Learning rates are tuned per method selected for speed, and the selected configuration is ran over five seeds with identical initialization and batch order across methods. In App.~\ref{app:benchmarks-protocol} we detail all experimental settings and explain the procedure for measuring FLOPs.

\begin{table}[h]
\caption{Reconstruction benchmarks.}
\centering
\small
\begin{adjustbox}{width=\linewidth}
\begin{tabular}{ll*{5}{r@{}l}}
\toprule
\textbf{Metric} & \textbf{Method} & \multicolumn{2}{c}{\textbf{MNIST}} & \multicolumn{2}{c}{\textbf{FMNIST}} & \multicolumn{2}{c}{\textbf{CIFAR}} & \multicolumn{2}{c}{\textbf{STL-10}} & \multicolumn{2}{c}{\textbf{Tiny ImageNet}} \\
\midrule
\multirow{6}{*}{\shortstack[l]{\textbf{MSE}\\($\times 10^{-3}$)}}
 & Standard PC & $4.41$&${}^{\pm 0.02}$ & $6.24$&${}^{\pm 0.01}$ & $5.36$&${}^{\pm 0.01}$ & $12.78$&${}^{\pm 0.01}$ & $15.04$&${}^{\pm 0.02}$ \\
 & iPC & $4.39$&${}^{\pm 0.01}$ & $6.29$&${}^{\pm 0.01}$ & $5.79$&${}^{\pm 0.01}$ & $12.75$&${}^{\pm 0.01}$ & $15.19$&${}^{\pm 0.01}$ \\
 & bPC & $4.44$&${}^{\pm 0.01}$ & $6.25$&${}^{\pm 0.01}$ & $5.35$&${}^{\pm 0.01}$ & $12.87$&${}^{\pm 0.01}$ & $15.08$&${}^{\pm 0.01}$ \\
 & ePC  & $4.41$&${}^{\pm 0.01}$ & $6.23$&${}^{\pm 0.01}$ & $5.44$&${}^{\pm 0.01}$ & $12.99$&${}^{\pm 0.01}$ & $14.94$&${}^{\pm 0.01}$ \\
 & PC-ALM  & $4.41$&${}^{\pm 0.01}$ & $6.22$&${}^{\pm 0.01}$ & $\mathbf{5.29}$&${}^{\pm 0.01}$ & $\mathbf{12.74}$&${}^{\pm 0.02}$ & $14.95$&${}^{\pm 0.00}$ \\
 & \textbf{Natural PC} & $\mathbf{4.38}$&${}^{\pm 0.02}$ & $\mathbf{6.21}$&${}^{\pm 0.01}$ & $5.33$&${}^{\pm 0.01}$ & $12.86$&${}^{\pm 0.01}$ & $\mathbf{14.92}$&${}^{\pm 0.01}$ \\
\midrule
\multirow{6}{*}{\textbf{Epochs}}
 & Standard PC & $26.6$&${}^{\pm 0.7}$ & $38.2$&${}^{\pm 1.3}$ & $38.2$&${}^{\pm 1.5}$ & $2.2$&${}^{\pm 0.0}$ & $2.4$&${}^{\pm 0.1}$ \\
 & iPC & $6.7$&${}^{\pm 0.1}$ & $7.0$&${}^{\pm 0.2}$ & $14.0$&${}^{\pm 2.1}$ & $1.2$&${}^{\pm 0.0}$ & $1.8$&${}^{\pm 0.1}$ \\
 & bPC & $127.0$&${}^{\pm 2.7}$ & $46.2$&${}^{\pm 1.9}$ & $26.2$&${}^{\pm 0.7}$ & $7.5$&${}^{\pm 0.0}$ & $3.8$&${}^{\pm 0.1}$ \\
 & ePC & $18.2$&${}^{\pm 0.8}$ & $37.5$&${}^{\pm 1.4}$ & $49.5$&${}^{\pm 1.5}$ & $3.4$&${}^{\pm 0.1}$ & $3.4$&${}^{\pm 0.1}$ \\
 & PC-ALM & $9.6$&${}^{\pm 0.4}$ & $16.2$&${}^{\pm 1.0}$ & $24.1$&${}^{\pm 0.4}$ & $1.7$&${}^{\pm 0.0}$ & $2.4$&${}^{\pm 0.1}$ \\
 & \textbf{Natural PC} & $\mathbf{1.4}$&${}^{\pm 0.1}$ & $\mathbf{1.6}$&${}^{\pm 0.1}$ & $\mathbf{1.3}$&${}^{\pm 0.2}$ & $\mathbf{0.08}$&${}^{\pm 0.0}$ & $\mathbf{0.08}$&${}^{\pm 0.0}$ \\
\midrule
\multirow{6}{*}{\textbf{W-updates}}
 & Standard PC & $1.3\text{k}$&${}^{\pm 35}$ & $1.9\text{k}$&${}^{\pm 65}$ & $1.9\text{k}$&${}^{\pm 76}$ & $1.1\text{k}$&${}^{\pm 20}$ & $1.2\text{k}$&${}^{\pm 44}$ \\
 & iPC & $10.1\text{k}$&${}^{\pm 147}$ & $10.6\text{k}$&${}^{\pm 278}$ & $21.0\text{k}$&${}^{\pm 3.2\text{k}}$ & $17.8\text{k}$&${}^{\pm 659}$ & $27.4\text{k}$&${}^{\pm 814}$ \\
 & bPC & $6.3\text{k}$&${}^{\pm 134}$ & $2.3\text{k}$&${}^{\pm 93}$ & $1.3\text{k}$&${}^{\pm 37}$ & $3.7\text{k}$&${}^{\pm 24}$ & $1.9\text{k}$&${}^{\pm 32}$ \\
 & ePC & $908$&${}^{\pm 41}$ & $1.8\text{k}$&${}^{\pm 72}$ & $2.4\text{k}$&${}^{\pm 77}$ & $1.7\text{k}$&${}^{\pm 49}$ & $1.6\text{k}$&${}^{\pm 59}$ \\
 & PC-ALM & $479$&${}^{\pm 18}$ & $808$&${}^{\pm 48}$ & $1.2\text{k}$&${}^{\pm 20}$ & $828$&${}^{\pm 20}$ & $1.2\text{k}$&${}^{\pm 34}$ \\
 & \textbf{Natural PC} & $\mathbf{72}$&${}^{\pm 6}$ & $\mathbf{78}$&${}^{\pm 3}$ & $\mathbf{66}$&${}^{\pm 10}$ & $\mathbf{39}$&${}^{\pm 2}$ & $\mathbf{38}$&${}^{\pm 2}$ \\
\midrule
\multirow{6}{*}{\textbf{TFLOP}}
 & Standard PC & $4.39$&${}^{\pm 0.1}$ & $6.31$&${}^{\pm 0.2}$ & $20.2$&${}^{\pm 0.8}$ & $98.7$&${}^{\pm 1.8}$ & $47.9$&${}^{\pm 1.7}$ \\
 & iPC & $2.20$&${}^{\pm 0.0}$ & $2.31$&${}^{\pm 0.1}$ & $14.6$&${}^{\pm 2.2}$ & $104$&${}^{\pm 4.0}$ & $72.0$&${}^{\pm 2.1}$ \\
 & bPC & $12.5$&${}^{\pm 0.3}$ & $4.55$&${}^{\pm 0.2}$ & $8.25$&${}^{\pm 0.2}$ & $198$&${}^{\pm 1.0}$ & $45.4$&${}^{\pm 0.8}$ \\
 & ePC & $3.02$&${}^{\pm 0.1}$ & $6.24$&${}^{\pm 0.2}$ & $26.4$&${}^{\pm 0.8}$ & $152$&${}^{\pm 4.0}$ & $67.6$&${}^{\pm 2.3}$ \\
 & PC-ALM & $1.58$&${}^{\pm 0.1}$ & $2.67$&${}^{\pm 0.2}$ & $12.7$&${}^{\pm 0.2}$ & $73.5$&${}^{\pm 1.8}$ & $48.8$&${}^{\pm 1.4}$ \\
 & \textbf{Natural PC} & $\mathbf{0.25}$&${}^{\pm 0.0}$ & $\mathbf{0.27}$&${}^{\pm 0.0}$ & $\mathbf{0.71}$&${}^{\pm 0.1}$ & $\mathbf{3.59}$&${}^{\pm 0.2}$ & $\mathbf{1.56}$&${}^{\pm 0.1}$ \\
\bottomrule
\end{tabular}
\end{adjustbox}
\label{tab:bench-results}
\end{table}

Our results are summarized in Tab.~\ref{tab:bench-results} with mean $\pm$ standard deviation over five random seeds. Natural PC reaches the lowest test error on most datasets, while requiring the fewest amount of weight updates, needing an order of magnitude fewer updates than standard PC. These efficiency gains translate to epochs as well, where Natural PC needs $237\times$ to $14\times$ fewer epochs than standard PC. Strikingly, on STL-10 and Tiny ImageNet, Natural PC converges after $0.08$ epochs, meaning well before the network has seen every training image. Natural PC also converges at the lowest cost of any method on every dataset, leading by about one order of magnitude. For consistency with the ImageNet experiments, we here present the results of models run with Nesterov Momentum on the inference steps, however, we also ran all models using regular, unaccelerated inference, which we report in App.~\ref{app:regular-inference}, Tab.~\ref{tab:bench-results-gd}. We do not find changes in method ordering or magnitude of differences in any of the metrics.

We also tested network-depth scaling, where Natural PC reaches the lowest error compared to existing PC methods at almost every depth and width, and is the only PC method that trains under ReLU at depth $32$ and beyond that we tested (App.~\ref{app:scaling}). Additionally, we tested the discriminative setting of PC in supervised classification \citep{whittington2017approximation,pinchetti2025benchmarkingpredictivecodingnetworks}. We observe that Natural PC gains $3\%$ to $19\%$ accuracy over standard PC and reaches the reference accuracy in $20\times$ to $12\times$ fewer epochs. Moreover, Natural PC also matches or exceeds BP's accuracy on all three tested datasets by $2.88$\% to $0.48$\% (App.~\ref{app:cls_results}, Tab.~\ref{tab:cls_full}), while requiring up to $5\times$ less weight updates.
Furthermore, Natural PC's classification accuracy also scales with depth. At $32$ and $64$ layers, Natural PC reaches $78\%$ and $51\%$ on MNIST, whereas AdamW diverges and standard PC falls to $35\%$ and $25\%$, respectively (App.~\ref{app:cls_results}, Tab.~\ref{tab:cls-depth-scaling}).

\section{Discussion}

Standard predictive coding wastes expensive inference by taking a single Euclidean step over the energy landscape, causing networks to require many inference phases before weights fit the equilibrium activities. Natural PC replaces the Euclidean learning step with a natural gradient step by adding a lateral circuit of interneurons that encode and implicitly invert the presynaptic correlations that define the natural gradient. We thereby improve the trade-off between inference and learning and show that biologically inspired local computation can compete with BP methods in both sample efficiency and computational cost in image reconstruction tasks. Beyond improving performance, the lateral circuit determines both the geometry and learning rate of weight updates. Whereas existing PC models and networks trained with BP require specifically chosen optimizers and learning rates, both emerge in Natural PC from the network's dynamics.

Although Natural PC is inherently not restricted to generative experiments, our method is not able to train discriminative networks at scale \citep{whittington2017approximation, kerjan2026trainingpredictivecodingnetwork}. We observe that in the generative setting, every layer predicts the high-dimensional activity of the layer below and thereby receives a dense prediction error, while in the discriminative setting, the target is a one-hot label that carries sparse error information to deeper layers after relaxation.
In line with previous observations \citep{zhao2017learning,pinchetti2025benchmarkingpredictivecodingnetworks,NEURIPS2025_7ea5ad70}, we notice that the top layer can sufficiently explain the target label, thereby blocking credit propagation to subsequent layers, which we coin the \textit{target absorption problem}.
To support discriminative PC at scale, we suggest combining the lateral circuit with mechanisms that improve prediction error propagation across layers \citep{qi2025trainingdeeperpredictivecoding,ICLR2026_319e084c,seely2026augmentedlagrangianpredictivecoding,kerjan2026trainingpredictivecodingnetwork}.
A second direction is to learn the precisions, as precision weighting has been related to natural gradient descent during inference \citep{millidge2021predictive,ofner2021predictivecodingprecisionnatural}.

Moreover, our results suggest a computational role for cortical interneurons in learning and motivate predictions about interneuron function, connectivity, and population size.
Lateral inhibition is often associated with decorrelation or whitening neural activity \citep{wiechert2010mechanisms,king2013inhibitory,chettih2019single,znamenskiy2024functional}.
We observe that whitening equilibrium activities provides a potential other way to implement natural gradient learning and, when tested, does improve standard PC(App.~\ref{app:implicit-natural-pc}). However, guiding synaptic learning directly without altering the representation, as is implemented by Natural PC, performs best.
Considering a cortical circuit that similarly regulates synaptic learning, leads us to two experimental predictions.
First, disrupting interneuron activity should impair learning efficiency without fully stopping the ability to learn.
We demonstrate this effect in our model by silencing the lateral circuit, which recovers standard PC updates while the model learns. We show that reactivating the circuit restores alignment with the natural gradient (App.~\ref{app:dissecting-circ}) and recovers the learning benefits.
Second, lateral connectivity should reflect neuronal co-activity even between neurons with dissimilar feature preferences, because the lateral learning rule tracks joint activity without requiring shared feature preferences.
Consistent with this distinction, lateral weights show little correlation with feature similarity in our models ($r\leq0.16$ across datasets), and paired recordings and connectivity mapping could test whether co-activity predicts lateral connectivity independently of feature similarity.
Finally, cortical inhibitory interneurons are substantially less numerous than principal neurons \citep{beaulieu1993numerical,sahara2012fraction}, raising the question of whether natural gradient learning requires equally sized neuronal populations.
Our analysis provides an indication that smaller populations may suffice, as the lateral weights reach an effective rank of only $17$ in layers of width $64$ (App.~\ref{app:dissecting-circ}), yet whether a smaller interneuron circuit preserves the learning benefits of Natural PC remains to be tested.

Natural PC brings predictive coding closer to a practical learning framework by enabling large-scale image reconstruction at computational costs competitive with BP. The observed gains in sample and compute efficiency enable local learning beyond small-scale settings which have constrained the field.  Our results open opportunities to investigate the proposed benefits of PC for online, continual, and sample-efficient learning, as well as to extend PC to larger vision, sequence, and language models.

\newpage
\appendix
\section{Layerwise Fisher Derivation}
\label{app:ng-derivation}

For the Gaussian conditional distribution in Eq.~\ref{eq:genmodel}, the score of the vectorized layer weights is

\begin{equation}
\nabla_{\operatorname{vec}(W_l)}\log p(z_{l-1}\mid z_l;W_l,\Pi_l)
=\phi(z_l)\otimes\Pi_l\varepsilon_l.
\label{eq:layerwise-score}
\end{equation}

The conditional residual covariance is $\mathbb{E}[\varepsilon_l\varepsilon_l^{\top}\mid z_l]=\Pi_l^{-1}$.
Taking the outer product of Eq.~\ref{eq:layerwise-score} and averaging over the model distribution gives
\begin{align}
\mathcal{F}_l
&=\mathbb{E}\!\left[
\bigl(\phi(z_l)\phi(z_l)^{\top}\bigr)
\otimes
\bigl(\Pi_l\varepsilon_l\varepsilon_l^{\top}\Pi_l\bigr)
\right] \\
&=\mathbb{E}\!\left[\phi(z_l)\phi(z_l)^{\top}\right]\otimes\Pi_l
=G_l\otimes\Pi_l.
\label{eq:layerwise-fisher-derivation}
\end{align}
The factorization is exact because the Gaussian residual covariance cancels one factor of $\Pi_l$.

\section{Lateral Circuit Equilibrium}
\label{app:lateral-circuit}

The lateral learning rule in Eq.~\ref{eq:lateral-circuit} averages outer products of settled presynaptic activities.
The equilibrium Gram matrix is

\begin{equation}
G_l^*\triangleq\mathbb{E}_{z_0\sim p_{\mathrm{data}}}\!\left[\phi(z_l^*)\phi(z_l^*)^{\top}\right].
\label{eq:equilibrium-gram-appendix}
\end{equation}

Taking the expectation over clamped data inputs therefore yields

\begin{equation}
\begin{aligned}
\tau_A\,\mathbb{E}[\dot A_l]
&=\mathbb{E}_{z_0\sim p_{\mathrm{data}}}\!\left[\phi(z_l^*)\phi(z_l^*)^{\top}-A_l\right]\\
&=G_l^*-A_l.
\end{aligned}
\label{eq:lateral-mean-dynamics}
\end{equation}

At convergence over training, the stationary mean of the lateral dynamics satisfies

\begin{equation}
A_l\longrightarrow G_l^*.
\label{eq:lateral-gram-convergence}
\end{equation}

For fixed positive-definite $A_l$, the equilibrium interneuron activity satisfies

\begin{equation}
0=\phi(z_l^*)-A_l a_l^*,
\qquad
a_l^*=A_l^{-1}\phi(z_l^*).
\label{eq:interneuron-fixed-point}
\end{equation}

Substituting Eq.~\ref{eq:interneuron-fixed-point} into the local weight update yields Eq.~\ref{eq:natural-pc-update}.

\section{Deriving the Newton, Gauss--Newton, and Fisher Identities}
\label{app:curvature-identities}

All derivatives below hold equilibrium activities and precision matrices fixed, and all data averages use $z_0\sim p_{\mathrm{data}}$.
Write $w_l=\operatorname{vec}(W_l)$ and define the prediction Jacobian by

\begin{equation}
J_l^*=\frac{\partial W_l\phi(z_l^*)}{\partial w_l}
=\phi(z_l^*)^\top\otimes I.
\end{equation}

The Gaussian negative log-likelihood in Eq.~\ref{eq:fixed-activity-objective} is

\begin{equation}
\bar{\mathcal E}_l(W_l)
=\tfrac12\mathbb E_{z_0}\!\left[\varepsilon_l^{*\top}\Pi_l\varepsilon_l^*\right]+c_l,
\qquad \varepsilon_l^*=z_{l-1}^*-W_l\phi(z_l^*),
\end{equation}

where $c_l$ is independent of $W_l$.
Since $\partial\varepsilon_l^*/\partial w_l=-J_l^*$ and $J_l^*$ is independent of the weights, direct differentiation gives

\begin{equation}
\begin{aligned}
\nabla_{w_l}\bar{\mathcal E}_l
&=-\mathbb E_{z_0}[J_l^{*\top}\Pi_l\varepsilon_l^*],\\
\mathcal H_l^*
=\nabla_{w_l}^2\bar{\mathcal E}_l
&=\mathbb E_{z_0}[J_l^{*\top}\Pi_lJ_l^*]
=\mathcal G_l^*.
\end{aligned}
\end{equation}

The middle expression is the Gauss--Newton definition for the precision-weighted least-squares objective.
For the Fisher, the layer output is sampled from the Gaussian conditional with $z_l^*$ fixed, giving $\varepsilon_l=z_{l-1}-W_l\phi(z_l^*)$ and $\mathbb E[\varepsilon_l\varepsilon_l^\top\mid z_l^*]=\Pi_l^{-1}$.
The score is $\nabla_{w_l}\log p=J_l^{*\top}\Pi_l\varepsilon_l$, so

\begin{equation}
\begin{aligned}
\mathcal F_l^*
&=\mathbb E_{z_0}\!\left[J_l^{*\top}\Pi_l
\mathbb E[\varepsilon_l\varepsilon_l^\top\mid z_l^*]
\Pi_lJ_l^*\right]\\
&=\mathbb E_{z_0}[J_l^{*\top}\Pi_lJ_l^*]
=\mathbb E_{z_0}[\phi(z_l^*)\phi(z_l^*)^\top]\otimes\Pi_l
=G_l^*\otimes\Pi_l.
\end{aligned}
\end{equation}

Thus $\mathcal H_l^*=\mathcal F_l^*=\mathcal G_l^*=G_l^*\otimes\Pi_l$.
The identities do not include derivatives through the equilibrium activities as the weights change.

\section{Exact Maximization at Fixed Equilibrium Activities}
\label{app:exact-maximization}

Hold the equilibrium activities and precisions fixed in Eq.~\ref{eq:fixed-activity-objective}, and assume $G_l^*$ and $\Pi_l$ are positive definite.
All expectations below use the same data distribution.
Differentiating the Gaussian objective gives

\begin{equation}
\nabla_{W_l}\bar{\mathcal E}_l
=\Pi_l\left(W_lG_l^*-\mathbb E[z_{l-1}^*\phi(z_l^*)^\top]\right).
\label{eq:fixed-activity-matrix-gradient}
\end{equation}

Setting the gradient to zero gives the unique minimizing weights, which also maximize the likelihood,

\begin{equation}
W_l^{\mathrm{opt}}
=\mathbb E[z_{l-1}^*\phi(z_l^*)^\top]G_l^{*-1},
\qquad
\nabla_{W_l}\bar{\mathcal E}_l
=\Pi_l(W_l-W_l^{\mathrm{opt}})G_l^*.
\label{eq:fixed-activity-optimum}
\end{equation}

The Fisher $G_l^*\otimes\Pi_l$ therefore gives the natural gradient update with unit learning rate,

\begin{equation}
\begin{aligned}
\Delta W_l^{\mathrm{NG}}
&=-\Pi_l^{-1}\nabla_{W_l}\bar{\mathcal E}_l\,G_l^{*-1}\\
&=-\Pi_l^{-1}\Pi_l(W_l-W_l^{\mathrm{opt}})G_l^*G_l^{*-1}
=W_l^{\mathrm{opt}}-W_l.
\end{aligned}
\label{eq:natural-exact-maximization}
\end{equation}

Adding the update to $W_l$ reaches $W_l^{\mathrm{opt}}$ and completes the likelihood maximization.
In contrast, a Euclidean update leaves the weight error

\begin{equation}
W_l-\eta\nabla_{W_l}\bar{\mathcal E}_l-W_l^{\mathrm{opt}}
=(W_l-W_l^{\mathrm{opt}})-\eta\Pi_l(W_l-W_l^{\mathrm{opt}})G_l^*,
\label{eq:euclidean-maximization-error}
\end{equation}

which generally cannot vanish for a single scalar learning rate when the remaining error spans directions with different curvature.
Standard PC does not perform additional updates to complete the maximization at fixed activities.
Exact one-step maximization requires the gradient and Fisher to use the same data average.
A stochastic gradient combined with a running Gram estimate does not generally satisfy the one-step identity.

\section{Learning Validation Experiments}
\label{app:adaptive-rate}

The experiments use 1000 activity-inference iterations per batch with an inference rate of $0.002$ and 2000 interneuron-relaxation iterations.
The lateral time constant is $\tau_A=20$ weight updates unless stated otherwise.
The reference statistics are computed on the first 10,000 training images, which also participate in training.
The reference set is therefore a larger training-set estimate of $G_l^*$, not a held-out set or an exact expectation over all 50,000 images.
Alignment is logged at every update through update 100 in the main run and every ten updates thereafter.
The time-constant sweep reports the mean alignment over the final ten logged measurements of 3000-update runs.
The figure uses one seed, and the spectral bands describe variation across directions, not uncertainty across independent runs.

The implementation uses precision-weighted prediction errors, so a code learning rate $\eta_l^{\mathrm{code}}$ corresponds to $\eta_l=\eta_l^{\mathrm{code}}/\sigma^2$ in Eq.~\ref{eq:effective-step}.
For standard PC, the same conversion applies when expressing the update using $A_l=I$ in the effective-step diagnostic.
The plotted standard PC alignment baseline uses $\eta^{\mathrm{code}}=10^{-2}$, and the effective-step baseline uses $\eta^{\mathrm{code}}=2\cdot10^{-2}$.
The latter rate was selected from the fixed-rate sweep.
Both methods also use weight decay with coefficient $0.03$ in the implementation.
The effective-step calculation isolates the likelihood update and does not include weight decay or changes in inferred activities.

For numerical stability, the circuit uses the regularized lateral matrix

\begin{equation}
A_{l,\rho}=A_l+\delta_l I,
\qquad
\delta_l=\max\!\left(\rho\,\frac{\operatorname{tr}(A_l)}{d_l},10^{-12}\right),
\qquad \rho=10^{-2}.
\label{eq:empirical-ridge}
\end{equation}

The reference Gram uses the same relative-ridge rule when inverted for alignment.
The weighted matrix inner product in Eq.~\ref{eq:alignment-cosine} is $\langle X,Y\rangle=\operatorname{tr}(XG_l^*Y^\top)$, with $\|X\|^2=\langle X,X\rangle$.
The activity-weighted cosine similarity used in Figure~\ref{fig:learning-theory}a is

\begin{equation}
\mathrm{align}(A_l,G_l^*)
=\frac{\operatorname{tr}(A_{l,\rho}^{-1}G_l^*G_{l,\rho}^{*-1})}
{\sqrt{\operatorname{tr}(A_{l,\rho}^{-1}G_l^*A_{l,\rho}^{-1})\,
\operatorname{tr}(G_{l,\rho}^{*-1}G_l^*G_{l,\rho}^{*-1})}}.
\label{eq:alignment}
\end{equation}

The similarity equals one when the inverse actions are positively proportional and measures directional agreement independently of their overall scale.
The effective-step calculation uses the eigenvalues of $A_{l,\rho}^{-1}G_l^*$, with the reference Gram left unregularized.

For evaluation, the main run uses an exponential weight average with rate $\kappa=0.05$.
Weight averaging is motivated by stochastic-approximation methods \citep{polyak1992acceleration}, although exponential averaging differs from the uniform averaging in the classical result.
After update $t$, the evaluation weights are

\begin{equation}
S_{l,t}=(1-\kappa)S_{l,t-1}+\kappa W_{l,t},
\qquad
\bar W_{l,t}=\frac{S_{l,t}}{1-(1-\kappa)^t},
\qquad S_{l,0}=0.
\label{eq:evaluation-weight-average}
\end{equation}

The bias correction removes the contribution of the initial zero accumulator.
Training always uses the current weights, and the evaluation average does not feed back into inference or learning during training.
The alignment run in panel (a) evaluates reference activities using the averaged weights, whereas panel (c) uses a separate run with unaveraged weights to measure the update geometry.
The distinction means that panel (a) compares the learned lateral matrix with the geometry of the evaluation model.

\section{Effective Step Analysis}
\label{app:effective-step}

Consider the likelihood update with equilibrium activities and the preconditioner held fixed, without weight decay.
Let $W_l^{\mathrm{opt}}$ minimize the reference objective and let $D_l=W_l-W_l^{\mathrm{opt}}$.
After cancelling the isotropic precision through the learning-rate convention above, the matrix error evolves as

\begin{equation}
D_l^+
=D_l\left(I-\eta_lG_l^*A_{l,\rho}^{-1}\right).
\label{eq:contraction}
\end{equation}

The eigenvalues of $G_l^*A_{l,\rho}^{-1}$ equal those of the symmetric matrix $A_{l,\rho}^{-1/2}G_l^*A_{l,\rho}^{-1/2}$.
In the corresponding transformed coordinates, a mode with eigenvalue $\lambda_i$ has error multiplier $1-\eta_l\lambda_i$.
The unnormalized effective step $s_{l,i}=\eta_l\lambda_i$ is therefore the fraction of the remaining distance covered along that mode.
A positive-curvature mode contracts for $0<s_{l,i}<2$, reaches the optimum for $s_{l,i}=1$, and overshoots for $s_{l,i}>1$.
Without the ridge, $A_l=G_l^*$ makes every eigenvalue one and recovers the exact unit step of Appendix~\ref{app:exact-maximization}.
A ridge or an imperfect Gram estimate generally prevents exact one-step maximization.
For a stochastic update, the spectral calculation describes the reference likelihood curvature and does not include the additional error from a minibatch gradient.

The eigenvalues are evaluated every ten updates, with medians and percentile bands as described in Figure~\ref{fig:learning-theory}c.
The plotted statistics are smoothed over five measurements in log space.

\newpage
\section{ImageNet Experiments}\label{app:imagenet-protocol}

\paragraph{Architecture.}
All PC methods use a bias-free VGG-10 decoder with $\tanh$ nonlinearities
and an 8,192-dimensional top code. A dense layer maps the code to a
$7\times7\times512$ tensor, followed by convolutional layers with spatial
resolutions and channel counts
$14\times14\times512$,
$14\times14\times512$,
$28\times28\times256$,
$28\times28\times256$,
$56\times56\times128$,
$56\times56\times128$,
$112\times112\times64$,
$112\times112\times64$, and
$224\times224\times3$.
The decoder contains 212.6M parameters.
The BP autoencoder uses the same decoder and an additional bottom-up
VGG-10 encoder that maps the image to the 8,192-dimensional code.

\paragraph{Data and evaluation.}
We train on the 1,281,167 training images of ImageNet-1k using random
$224\times224$ crops and horizontal flips, with pixels in $[0,1]$.
Each update uses 100 images, and each run uses one GH200 on CSCS Clariden.
We evaluate on 5,000 validation images, with five images per class.
For PC methods, the reported metric is the pixel MSE of the image decoded
from the inferred top code. For BP, we report the MSE of the
autoencoder reconstruction.
PC methods use 300 inference iterations at evaluation.
Evaluation uses a running weight average with memory $M=10$ updates.
All PC methods use a Gaussian prior on the top code with standard
deviation 4.

\paragraph{Hyperparameter selection.}
For methods with a learning rate, we use linear warm-up over the first
5,000 training images, corresponding to 50 updates, followed by cosine
decay to zero at the end of the run.
Within each sweep, we select the run with the lowest MSE at the final
evaluation.
The preliminary BP sweeps described below use a smaller code width and
larger batch size than the final experiments where indicated.
We distinguish these preliminary results from the settings used in the
final runs.

\paragraph{Natural PC.}
Training inference uses 30 iterations of restarted Nesterov acceleration
per batch, with a layerwise inference rate of $0.7$ divided by the
curvature bound.
The M-step uses recursive least squares with a unit step and memory
$M=10$ updates for the lateral matrices, step-size estimates, noise
variances, and weight average.
Weight decay is 0.03, scaled by the step size.
We choose the inference settings, memory, and weight decay on a
100,000-image subset, selecting the inference settings for the lowest
loss per second.
Natural PC does not need learning-rate sweeps.
The run uses a 10-epoch schedule and is stopped at update 9,420.
The reported checkpoint is evaluated after 501,200 training images,
with reconstruction MSE $0.002303$.

\paragraph{BP autoencoder.}
We train the autoencoder with AdamW, Muon, SOAP, and SGD.
AdamW, Muon, and SOAP use decoupled weight decay, whereas SGD uses
coupled $L_2$ regularization.
SOAP refreshes its eigenbasis every 10 updates.

For AdamW, we sweep learning rates
$\{3\times10^{-5},10^{-4},3\times10^{-4}\}$ at weight decay 0.1.
After one million training images, these settings yield MSEs
$\{0.00494,0.00419,0.00848\}$, respectively.
We select learning rate $10^{-4}$ and weight decay 0.1.
The final 25-epoch run reaches MSE $0.00181$.

For Muon, a preliminary sweep uses code width 1,024 and batch size 256.
We sweep learning rates
$\{0.003,0.005,0.01,0.02,0.05\}$ at weight decay 0.03,
obtaining MSEs
$\{0.0080,0.0073,0.0066,0.0077,0.0130\}$
after 400,000 images.
At learning rate 0.01, weight decays $\{0.03,0.1,0.3\}$ yield
MSEs $\{0.0066,0.0062,0.0065\}$.
We subsequently refine the learning rate over the range 0.005 to 0.04
at code width 8,192 using 100,000 images.
The final setting is learning rate 0.0075 and weight decay 0.1;
the 10-epoch run reaches MSE $0.00278$.

For SOAP, the preliminary sweep uses code width 1,024 and batch size 256.
Learning rates
$\{10^{-4},3\times10^{-4},10^{-3},3\times10^{-3}\}$
at weight decay 0.03 yield MSEs
$\{0.0186,0.0108,0.0155,0.0182\}$
after 400,000 images.
The final 10-epoch run uses learning rate $10^{-3}$ and weight decay
0.1, reaching MSE $0.00650$.

For SGD, the preliminary sweep also uses code width 1,024 and batch
size 256.
Learning rates $\{0.03,0.1,0.3\}$ without $L_2$ regularization yield
MSEs $\{0.0273,0.0232,0.0723\}$ after 400,000 images.
The final 10-epoch run uses learning rate 0.3 and $L_2$ coefficient
$5\times10^{-4}$, reaching MSE $0.00587$.

Earlier preliminary sweeps also included Shampoo and Lion.
Lion remained near the mean-image reconstruction at every tested
learning rate, and Shampoo was not retained in the final comparison.

\paragraph{PC baselines.}
We compare standard PC, iPC, ePC, bPC, and PC-ALM.
All five methods use fixed unit noise variances at every level and
AdamW on their local weight gradients without weight decay.
They share Natural PC's 30-iteration restarted Nesterov inference
procedure, top-code prior, and evaluation protocol.
Each run is trained for one epoch, corresponding to 12,812 updates,
except for two iPC configurations stopped early after divergence.

For standard PC, learning rates
$\{10^{-5},3\times10^{-5},10^{-4},3\times10^{-4}\}$
yield final MSEs $\{0.0305,0.0210,0.0175,0.0379\}$.
We select learning rate $10^{-4}$.

For iPC, weights are updated at every inference iteration, and the
specified learning rate applies to each such update.
Learning rates $\{10^{-7},3\times10^{-7},10^{-6}\}$ yield final MSEs
$\{0.151,0.332,0.675\}$, with minimum MSEs during training of
$\{0.0325,0.0278,0.0257\}$.
Runs at $3\times10^{-6}$ and $10^{-5}$ are stopped after 50,000
images following divergence, with MSEs 0.111 and 0.165.
We select learning rate $10^{-7}$.

For ePC, inference is performed in error coordinates at rate 0.7.
Learning rates $\{10^{-5},3\times10^{-5},10^{-4}\}$ yield final MSEs
$\{0.0674,0.0686,0.0688\}$.
The run at $3\times10^{-4}$ produces NaNs after 200,000 images.
We select learning rate $10^{-5}$.

For bPC, the bottom-up error variance is $\kappa=1$, and the
bottom-up learning rate is tied to the decoder learning rate.
Learning rates $\{10^{-5},3\times10^{-5},10^{-4}\}$ yield final MSEs
$\{0.171,0.318,0.243\}$, with minimum MSEs during training of
$\{0.0397,0.0412,0.0405\}$.
We select learning rate $10^{-5}$.

For PC-ALM, we take one dual step per inference iteration without
ramping the dual rate.
The pairs of weight learning rate and dual rate
$(\eta,\alpha)\in
\{(10^{-4},0.03),(10^{-4},0.1),(10^{-4},0.3),
(3\times10^{-5},0.1)\}$
yield final MSEs $\{0.0194,0.0531,0.195,0.0270\}$.
We select $(\eta,\alpha)=(10^{-4},0.03)$.
The published default $\alpha=1$ is unstable with the
30-iteration Nesterov inference procedure. Preliminary stability
tests support $\alpha\leq0.3$.

\paragraph{Baseline behavior and runtime.}
No PC baseline reaches MSE 0.012, the easiest reconstruction target
in Fig.~3.
Standard PC and PC-ALM exhibit transient increases in MSE to between
0.1 and 0.8 over approximately 10,000 to 200,000 training images,
before recovering as the learning rate decays.
Both iPC and bPC reach their lowest MSE within the first 100,000
images and subsequently deteriorate at every tested learning rate.
Meanwhile, ePC remains near the mean-image MSE of approximately 0.07.
The selected standard PC, iPC, ePC, bPC, and PC-ALM runs take
0.35, 2.28, 0.52, 1.98, and 0.41 seconds per batch update,
respectively.

\paragraph{Classification.}
For reference, a backprop VGG-10 classifier on the same data reaches $32.1\%$ top-1 accuracy after $10$ epochs at batch size $100$ with a flat head, and $30.8\%$ after $20$ epochs at batch size $50$ with a global-average-pooling head. A discriminative Natural PC classifier with the construction of Appendix~\ref{app:cls_results} did not learn at this scale, so we report ImageNet results for the generative model only.

\newpage
\section{Benchmark Protocol}\label{app:benchmarks-protocol}
\paragraph{Model and data.}
All methods use a generative model with three hidden layers of 128 units
and $\tanh$ nonlinearities. We evaluate on MNIST, FashionMNIST,
CIFAR-10, STL-10, and Tiny ImageNet, with images flattened and pixels
scaled to $[0,1]$. Training uses the official training splits of MNIST,
FashionMNIST, CIFAR-10, and Tiny ImageNet, and the unlabeled split of
STL-10. Evaluation uses 512 held-out images from the official test splits
of MNIST, FashionMNIST, and CIFAR-10, the labeled training split of
STL-10, and the validation split of Tiny ImageNet.
Mini-batches contain 200 images.

\paragraph{Natural PC.}
Natural PC uses the implementation of the ImageNet recipe (App~.\ref{app:imagenet-protocol}),
including 30 restarted-Nesterov inference iterations per mini-batch.
The lateral circuit implements a unit-step recursive least-squares
update, with memory $M=10$ updates for its running estimates and weight
average. Evaluation uses 300 inference iterations and the running weight
average with averaging rate $1/10$.

\paragraph{Standard PC in Table~1.}
Standard PC takes one plain gradient step on the weights after the
30-step inference phase, using a constant learning rate.
On MNIST, FashionMNIST, and CIFAR-10, decoupled weight decay is
$0.03\eta$ per update, where $\eta$ is the weight learning rate.
On STL-10 and Tiny ImageNet, the decay is fixed at
$3\times10^{-5}$ per update.

\paragraph{iPC in Table~1.}
iPC performs 30 inner steps per mini-batch, each consisting of one
restarted-Nesterov step on the activities followed by one AdamW
step on the weights.
The inference step sizes are recomputed from the current weights at
every inner step, and the weight learning rate remains constant.
Each mini-batch therefore counts as 30 weight updates.
Training uses a unit Gaussian prior on every hidden latent and fixed decay of $1.5\times10^{-6}$ per weight
update.

\paragraph{Additional PC variants.}
PC with AdamW, ePC, and PC-ALM share Natural PC's inference and
evaluation procedure, including 300 evaluation iterations and a running
weight average with rate $1/10$.
They use a Gaussian top-code prior with standard deviation 4 and
AdamW with $(\beta_1,\beta_2)=(0.9,0.999)$ and no weight decay.
The learning rate increases linearly over the first 5,000 training
images and remains constant thereafter.

PC with AdamW applies AdamW to the local PC weight gradient after
inference. In ePC, inference instead operates in error coordinates,
\[
z_l = W_{l+1}\phi(z_{l+1}) + e_l,
\]
with inference rate 0.7 divided by the curvature bound of the
error-coordinate energy.
PC-ALM performs inference on the augmented Lagrangian, taking one
dual-ascent step on the hidden-level multipliers per inference
iteration, with dual rate 0.01 from the start.

\paragraph{bPC implementation.}
bPC augments the top-down generative model with a separate set of
bottom-up prediction weights. Each hidden layer is therefore predicted
both from the layer above and from the layer below. With the image
clamped, inference adjusts the hidden activities to jointly minimize
the bottom-up and top-down prediction errors. Both sets of weights
are then updated using their corresponding prediction errors.
Intuitively, the bottom-up pathway enforces hidden activities to
remain predictable from the sensory input, while the top-down pathway
learns to reconstruct it. We use a dense implementation of the
unsupervised MLP configuration from the authors' repository.

\paragraph{Learning-rate tuning}
We sweep learning rates from $3\times10^{-4}$ to $0.1$ for
standard PC, $1.5\times10^{-5}$ to $0.14$ for iPC,
$10^{-4}$ to $0.03$ for PC with AdamW and PC-ALM,
and $10^{-4}$ to $0.09$ for ePC, using dataset-specific grids.
For bPC, we sweep parameter learning rates
$\eta_p\in\{10^{-5},3\times10^{-5},10^{-4},3\times10^{-4},10^{-3}\}$
and activity inference rates
$\eta_x\in\{10^{-3},3\times10^{-3},10^{-2},3\times10^{-2},0.1\}$.
We additionally test $\eta_p=3\times10^{-3}$ at $\eta_x=10^{-2}$
on FashionMNIST and CIFAR-10, and at $\eta_x=3\times10^{-3}$
on Tiny ImageNet.
Table~\ref{tab:selected-lrs} lists the selected rates.

\begin{table}[ht]
\centering
\small
\caption{Selected learning rates per method and dataset.
The iPC rate applies to each inner weight update.
For bPC, separate rows report the parameter learning rate $\eta_p$
and activity inference rate $\eta_x$.
Natural PC requires no weight learning rate.}
\label{tab:selected-lrs}
\begin{tabular}{lccccc}
\hline
Method & MNIST & FMNIST & CIFAR-10 & STL-10 & Tiny ImageNet \\
\hline
Natural PC & --- & --- & --- & --- & --- \\
Standard PC & 0.04 & 0.03 & 0.01 & 0.03 & 0.03 \\
iPC & 0.005 & 0.005 & 0.01 & 0.045 & 0.005 \\
PC (AdamW) & 0.01 & 0.01 & 0.01 & 0.01 & 0.003 \\
ePC & 0.03 & 0.03 & 0.03 & 0.01 & 0.01 \\
PC-ALM & 0.01 & 0.01 & 0.01 & 0.01 & 0.003 \\
bPC ($\eta_p$) & 0.0003 & 0.001 & 0.001 & 0.0003 & 0.001 \\
bPC ($\eta_x$) & 0.01 & 0.01 & 0.01 & 0.003 & 0.003 \\
\hline
\end{tabular}
\end{table}

\begin{table}[!htbp]
\centering
\caption{Performance and cost per method at two widths (layers $[D,w,w,128]$), following the protocol of Appendix~\ref{app:benchmarks-protocol}. Every baseline uses the learning rate that reaches the plateau in the fewest weight updates among grid points within $0.5\%$ of the best mean MSE with all seeds converged. MSE is the lowest training error of a run, and epochs, W-updates and TFLOP are counted to the first evaluation within $2\%$ of that minimum. \emph{Updates to Std.\ PC plateau} counts the weight updates until the training error first reaches $1.02\times$ the mean Standard PC MSE of the same cell. Mean $\pm$ standard deviation over $5$ seeds.}
\vspace{0.6em}
\small
\begin{adjustbox}{width=\linewidth}
\begin{tabular}{llcccc}
\toprule
\multirow{2}{*}{\textbf{Metric}} &
\multirow{2}{*}{\textbf{Method}} &
\multicolumn{2}{c}{\textbf{STL-10}} &
\multicolumn{2}{c}{\textbf{Tiny ImageNet}} \\
\cmidrule(lr){3-4}\cmidrule(lr){5-6}
& & \textbf{Width 128} & \textbf{Width 256} & \textbf{Width 128} & \textbf{Width 256} \\
\midrule
\multirow{4}{*}{\shortstack[l]{\textbf{MSE}\\($\times 10^{-3}$)}}
 & Standard PC & $12.78^{\pm 0.01}$ & $9.48^{\pm 0.01}$ & $15.04^{\pm 0.02}$ & $11.67^{\pm 0.02}$ \\
 & iPC & $12.75^{\pm 0.01}$ & $11.42^{\pm 0.02}$ & $15.19^{\pm 0.01}$ & $13.98^{\pm 0.01}$ \\
& bPC & $12.87^{\pm 0.01}$ & $\mathbf{9.44}^{\pm 0.01}$ & $15.08^{\pm 0.01}$ & $11.69^{\pm 0.00}$ \\
   & ePC & $12.99^{\pm 0.01}$ & $9.47^{\pm 0.00}$ & $14.94^{\pm 0.01}$ & $11.71^{\pm 0.00}$ \\
 & PC-ALM & $\mathbf{12.74}^{\pm 0.02}$ & $9.47^{\pm 0.01}$ & $14.95^{\pm 0.00}$ & $11.71^{\pm 0.01}$ \\
 & \textbf{Natural PC} & $12.86^{\pm 0.01}$ & $\mathbf{9.44}^{\pm 0.02}$ & $\mathbf{14.92}^{\pm 0.01}$ & $\mathbf{11.69}^{\pm 0.02}$ \\

\midrule
\multirow{4}{*}{\textbf{Epochs}}
 & Standard PC & $2.22^{\pm 0.04}$ & $6.91^{\pm 0.09}$ & $2.40^{\pm 0.09}$ & $7.07^{\pm 0.38}$ \\
 & iPC & $1.19^{\pm 0.04}$ & $2.78^{\pm 0.12}$ & $1.82^{\pm 0.05}$ & $1.40^{\pm 0.04}$ \\
& bPC & $7.50^{\pm 0.05}$ & $7.50^{\pm 0.04}$ & $3.82^{\pm 0.06}$ & $6.51^{\pm 0.05}$ \\
 & ePC  & $3.40^{\pm 0.10}$ & $6.29^{\pm 0.14}$ & $3.37^{\pm 0.12}$ & $6.03^{\pm 0.23}$ \\
 & PC-ALM & $1.66^{\pm 0.04}$ & $3.35^{\pm 0.07}$ & $2.45^{\pm 0.07}$ & $3.30^{\pm 0.09}$ \\
 & \textbf{Natural PC} & $\mathbf{0.079}^{\pm 0.004}$ & $\mathbf{0.083}^{\pm 0.003}$ & $\mathbf{0.076}^{\pm 0.004}$ & $\mathbf{0.082}^{\pm 0.007}$ \\

\midrule
\multirow{4}{*}{\textbf{W-updates}}
 & Standard PC & $1{,}112^{\pm 20}$ & $3{,}456^{\pm 46}$ & $1{,}200^{\pm 44}$ & $3{,}536^{\pm 189}$ \\
 & iPC & $17.8\text{k}^{\pm 0.7\text{k}}$ & $41.6\text{k}^{\pm 1.8\text{k}}$ & $27.4\text{k}^{\pm 0.8\text{k}}$ & $21.0\text{k}^{\pm 0.7\text{k}}$ \\
& bPC & $3{,}748^{\pm 24}$ & $3{,}748^{\pm 20}$ & $1{,}912^{\pm 32}$ & $3{,}256^{\pm 27}$ \\
 & ePC  & $1{,}700^{\pm 49}$ & $3{,}144^{\pm 72}$ & $1{,}684^{\pm 59}$ & $3{,}016^{\pm 116}$ \\
 & PC-ALM & $828^{\pm 20}$ & $1{,}676^{\pm 34}$ & $1{,}224^{\pm 34}$ & $1{,}652^{\pm 45}$ \\
 & \textbf{Natural PC} & $\mathbf{39}^{\pm 2}$ & $\mathbf{42}^{\pm 1}$ & $\mathbf{38}^{\pm 2}$ & $\mathbf{41}^{\pm 4}$ \\

\midrule
\multirow{4}{*}{\shortstack[l]{\textbf{Updates to}\\\textbf{Std. PC plateau}}}
 & Standard PC & $1{,}212^{\pm 16}$ & $3{,}664^{\pm 86}$ & $1{,}288^{\pm 37}$ & $3{,}776^{\pm 164}$ \\
 & iPC & $17.5\text{k}^{\pm 0.6\text{k}}$ & never & $85.2\text{k}^{\pm 8.1\text{k}}$ & never \\
& bPC & $3{,}728^{\pm 30}$ & $3{,}752^{\pm 32}$ & $2{,}184^{\pm 50}$ & $3{,}484^{\pm 34}$ \\
 & ePC & $1{,}660^{\pm 33}$ & $3{,}080^{\pm 59}$ & $1{,}656^{\pm 37}$ & $3{,}000^{\pm 110}$ \\
 & PC-ALM & $844^{\pm 15}$ & $1{,}624^{\pm 50}$ & $1{,}200^{\pm 36}$ & $1{,}632^{\pm 56}$ \\
 & \textbf{Natural PC} & $\mathbf{19}^{\pm 1}$ & $\mathbf{40}^{\pm 3}$ & $\mathbf{32}^{\pm 1}$ & $\mathbf{38}^{\pm 3}$ \\

\midrule
\multirow{4}{*}{\textbf{TFLOP}}
 & Standard PC & $98.7^{\pm 1.8}$ & $616^{\pm 8}$ & $47.9^{\pm 1.7}$ & $285^{\pm 15}$ \\
 & iPC & $104^{\pm 4}$ & $490^{\pm 21}$ & $72.0^{\pm 2.1}$ & $112^{\pm 3}$ \\
& bPC & $198^{\pm 1}$ & $398^{\pm 2}$ & $45.4^{\pm 0.8}$ & $156^{\pm 1}$ \\
 & ePC & $152^{\pm 4}$ & $564^{\pm 13}$ & $67.6^{\pm 2.3}$ & $245^{\pm 9}$ \\
 & PC-ALM (AdamW) & $73.5^{\pm 1.8}$ & $299^{\pm 6}$ & $48.8^{\pm 1.4}$ & $133^{\pm 4}$ \\
 & \textbf{Natural PC} & $\mathbf{3.59}^{\pm 0.17}$ & $\mathbf{7.69}^{\pm 0.28}$ & $\mathbf{1.56}^{\pm 0.09}$ & $\mathbf{3.43}^{\pm 0.30}$ \\

\bottomrule
\end{tabular}
\end{adjustbox}
\label{tab:scaling}
\end{table}

\subsection{Depth and Width Ablation in Reconstruction}
\label{app:scaling}

We also report depth and width scaling results on the exact same experimental setup like in the main text in Tab.~\ref{scaling_recon}, detailed in App.~\ref{app:benchmarks-protocol}. Diverged indicates that at least 2 of 5 seeds diverged during training. Mean $\pm$ std are reported over 5 seeds. The learning rate was individually tuned for each method.

\begin{table}[!htbp]
\centering
\caption{Reconstruction Error by Depth, Width ($W=128$), Activation Function and Method}
\vspace{0.6em}
\small
\begin{adjustbox}{width=\linewidth}
\begin{tabular}{llcccccc}
\toprule
 & & \multicolumn{3}{c}{\textbf{tanh}} & \multicolumn{3}{c}{\textbf{ReLU}} \\
\cmidrule(lr){3-5}\cmidrule(lr){6-8}
\textbf{Depth} & \textbf{Method} & MNIST & FMNIST & CIFAR & MNIST & FMNIST & CIFAR \\
\midrule
 \multirow{6}{*}{4} & Standard PC & $5.33^{\pm 0.02}$ & $7.83^{\pm 0.03}$ & $5.82^{\pm 0.00}$ & $5.36^{\pm 0.08}$ & $7.93^{\pm 0.07}$ & $11.33^{\pm 1.22}$ \\
  & iPC & $14.00^{\pm 0.04}$ & $24.66^{\pm 0.16}$ & $61.13^{\pm 0.01}$ & $17.33^{\pm 0.16}$ & $24.10^{\pm 0.10}$ & $29.27^{\pm 0.30}$ \\
  & \textbf{Natural PC} & $4.09^{\pm 0.03}$ & $6.74^{\pm 0.13}$ & $5.55^{\pm 0.20}$ & $4.84^{\pm 0.03}$ & $\mathbf{6.64}^{\pm 0.19}$ & $6.66^{\pm 0.15}$ \\
  & Implicit Natural PC & $4.98^{\pm 0.01}$ & $7.59^{\pm 0.26}$ & $5.94^{\pm 0.04}$ & $8.26^{\pm 0.04}$ & $11.02^{\pm 0.04}$ & \emph{diverged} \\
  & Backprop (AdamW) & $4.08^{\pm 0.00}$ & $6.12^{\pm 0.00}$ & $5.31^{\pm 0.01}$ & $\mathbf{4.59}^{\pm 0.01}$ & $6.75^{\pm 0.03}$ & $\mathbf{6.62}^{\pm 0.33}$ \\
      & Backprop (SGD) & $4.06^{\pm 0.01}$ & \emph{diverged} & \emph{diverged} & $5.29^{\pm 0.08}$ & $7.28^{\pm 0.09}$ & $7.25^{\pm 0.30}$ \\
  & Backprop (L-BFGS) & $\mathbf{4.02}^{\pm 0.01}$ & $\mathbf{6.07}^{\pm 0.00}$ & $\mathbf{5.29}^{\pm 0.00}$ & \emph{diverged} & $17.81^{\pm 4.01}$ & $23.04^{\pm 0.61}$ \\
\midrule
 \multirow{6}{*}{8} & Standard PC & $5.33^{\pm 0.02}$ & $7.93^{\pm 0.07}$ & $6.17^{\pm 0.02}$ & $5.24^{\pm 0.07}$ & $7.86^{\pm 0.06}$ & \emph{diverged} \\
  & iPC & $13.80^{\pm 0.03}$ & $21.90^{\pm 0.09}$ & $61.13^{\pm 0.01}$ & $18.19^{\pm 0.16}$ & $25.83^{\pm 0.09}$ & $31.42^{\pm 0.22}$ \\
  & \textbf{Natural PC} & $4.12^{\pm 0.11}$ & $6.76^{\pm 0.27}$ & $\mathbf{5.27}^{\pm 0.09}$ & $5.19^{\pm 0.08}$ & $7.12^{\pm 0.23}$ & $\mathbf{6.86}^{\pm 0.31}$ \\
  & Implicit Natural PC & $4.93^{\pm 0.01}$ & $6.77^{\pm 0.02}$ & $5.79^{\pm 0.06}$ & $8.28^{\pm 0.03}$ & \emph{diverged} & \emph{diverged} \\
  & Backprop (AdamW) & $4.08^{\pm 0.00}$ & $6.13^{\pm 0.01}$ & $5.31^{\pm 0.01}$ & $\mathbf{4.98}^{\pm 0.04}$ & $\mathbf{7.03}^{\pm 0.04}$ & $6.90^{\pm 0.40}$ \\
      & Backprop (SGD) & $4.08^{\pm 0.01}$ & \emph{diverged} & \emph{diverged} & $5.78^{\pm 0.08}$ & $8.18^{\pm 0.15}$ & $7.51^{\pm 0.13}$ \\
  & Backprop (L-BFGS) & $\mathbf{4.03}^{\pm 0.00}$ & $\mathbf{6.07}^{\pm 0.00}$ & $5.28^{\pm 0.01}$ & $22.94^{\pm 7.61}$ & $16.09^{\pm 0.08}$ & $20.51^{\pm 0.77}$ \\
\midrule
 \multirow{6}{*}{32} & Standard PC & $5.12^{\pm 0.02}$ & $7.61^{\pm 0.07}$ & $6.04^{\pm 0.02}$ & \emph{diverged} & \emph{diverged} & \emph{diverged} \\
  & iPC & $13.21^{\pm 0.02}$ & $22.21^{\pm 0.08}$ & $61.12^{\pm 0.00}$ & $17.81^{\pm 0.14}$ & $25.46^{\pm 0.10}$ & $31.43^{\pm 0.21}$ \\
  & \textbf{Natural PC} & $\mathbf{4.02}^{\pm 0.18}$ & $6.13^{\pm 0.14}$ & $5.39^{\pm 0.22}$ & $5.03^{\pm 0.09}$ & $7.12^{\pm 0.12}$ & $6.56^{\pm 0.47}$ \\
  & Implicit Natural PC & $4.74^{\pm 0.01}$ & $7.11^{\pm 0.44}$ & $5.80^{\pm 0.05}$ & $7.94^{\pm 0.04}$ & \emph{diverged} & \emph{diverged} \\
  & Backprop (AdamW) & $4.08^{\pm 0.00}$ & $6.12^{\pm 0.01}$ & $5.33^{\pm 0.01}$ & $\mathbf{4.92}^{\pm 0.05}$ & $\mathbf{6.96}^{\pm 0.01}$ & $\mathbf{6.37}^{\pm 0.32}$ \\
      & Backprop (SGD) & $4.08^{\pm 0.01}$ & \emph{diverged} & \emph{diverged} & $5.75^{\pm 0.09}$ & \emph{diverged}& $7.47^{\pm 0.09}$ \\
  & Backprop (L-BFGS) & $4.04^{\pm 0.00}$ & $\mathbf{6.09}^{\pm 0.00}$ & $\mathbf{5.30}^{\pm 0.01}$ & $19.06^{\pm 3.56}$ & $16.28^{\pm 0.32}$ & $20.29^{\pm 0.89}$ \\
\midrule
 \multirow{6}{*}{64} & Standard PC & $5.12^{\pm 0.02}$ & $7.61^{\pm 0.07}$ & $6.04^{\pm 0.02}$ & \emph{diverged} & \emph{diverged} & \emph{diverged} \\
  & iPC & $13.21^{\pm 0.02}$ & $22.22^{\pm 0.07}$ & $61.12^{\pm 0.00}$ & $17.81^{\pm 0.14}$ & $25.46^{\pm 0.10}$ & $31.44^{\pm 0.21}$ \\
  & \textbf{Natural PC} & $\mathbf{3.94}^{\pm 0.08}$ & $6.29^{\pm 0.24}$ & $5.45^{\pm 0.21}$ & $5.09^{\pm 0.10}$ & $\mathbf{6.83}^{\pm 0.18}$ & $\mathbf{6.51}^{\pm 0.19}$ \\
  & Implicit Natural PC & $4.74^{\pm 0.01}$ & $7.08^{\pm 0.41}$ & $5.81^{\pm 0.05}$ & $7.93^{\pm 0.05}$ & \emph{diverged} & \emph{diverged} \\
  & Backprop (AdamW) & $4.08^{\pm 0.01}$ & $6.33^{\pm 0.01}$ & $\mathbf{5.39}^{\pm 0.02}$ & $\mathbf{4.91}^{\pm 0.01}$ & $6.94^{\pm 0.01}$ & $6.54^{\pm 0.50}$ \\
 & Backprop (SGD) & $4.78^{\pm 0.11}$ & $7.14^{\pm 0.13}$ & $6.09^{\pm 0.08}$ & $5.74^{\pm 0.05}$ & $8.11^{\pm 0.08}$ & $7.52^{\pm 0.06}$ \\
  & Backprop (L-BFGS) & $4.05^{\pm 0.00}$ & $\mathbf{6.12}^{\pm 0.01}$ &\emph{diverged} & $17.58^{\pm 0.51}$ & $16.08^{\pm 0.05}$ & $19.90^{\pm 0.52}$ \\
\bottomrule
\end{tabular}
\end{adjustbox}
\label{scaling_recon}
\end{table}

\newpage

\subsection{Results with regular inference}
\label{app:regular-inference}
Tables~\ref{tab:bench-results-gd} and~\ref{tab:scaling-gd} repeat the benchmarks with regular inference: $1{,}000$ gradient-descent steps at a searched lr, for training.
Everything else follows App.~\ref{app:benchmarks-protocol}, except for the following.
iPC runs $T=20$ inner steps per mini-batch at inference rate $0.01$ on MNIST, FashionMNIST and CIFAR-10 and $0.002$ on STL-10 and Tiny ImageNet.
Backprop unrolls all $1{,}000$ steps.
Natural PC uses the earlier form of the lateral rule: it tracks the batch Gram with an exponential moving average at rate $\alpha_A=0.05$, applies $A_l^{-1}$ by $2{,}000$ interneuron relaxation steps on the ridge-regularized matrix ($\rho=10^{-2}$), sets the rate of each layer from the largest eigenvalue of $A_{l,\rho}^{-1}\hat G_l$, and is evaluated on an exponential average of its weights at rate $0.05$.
Each model is evaluated with the inference it was trained with. Table~\ref{tab:selected-lrs} lists the learning rates.

\begin{table}[ht]
\caption{Reconstruction benchmarks with the regular inference ($1{,}000$ gradient-descent steps at rate $0.002$ for training and evaluation), as Table~\ref{tab:bench-results}.}
\centering
\small
\begin{adjustbox}{width=\linewidth}
\begin{tabular}{llccccc}
\toprule
\textbf{Metric} & \textbf{Method} & \textbf{MNIST} & \textbf{FMNIST} & \textbf{CIFAR} & \textbf{STL-10} & \textbf{Tiny ImageNet} \\
\midrule
\multirow{6}{*}{\shortstack[l]{\textbf{MSE}\\($\times 10^{-3}$)}}

& Standard PC & $4.52^{\pm 0.01}$ & $6.37^{\pm 0.01}$ & $5.51^{\pm 0.01}$ & $13.41^{\pm 0.24}$ & $15.26^{\pm 0.03}$ \\
 & iPC & $4.58^{\pm 0.01}$ & $6.57^{\pm 0.01}$ & $6.35^{\pm 0.02}$ & $14.90^{\pm 0.04}$ & $15.49^{\pm 0.01}$ \\
  & ePC  & $4.87^{\pm 0.02}$ & $6.77^{\pm 0.14}$ & $6.13^{\pm 0.08}$ & $13.96^{\pm 0.06}$ & $15.79^{\pm 0.48}$ \\
 & PC-ALM & $4.41^{\pm 0.00}$ & $\mathbf{6.21}^{\pm 0.02}$ & $5.37^{\pm 0.03}$ & $13.35^{\pm 0.11}$ & $15.20^{\pm 0.03}$ \\
 & bPC & $4.44^{\pm 0.01}$ & $6.23^{\pm 0.01}$ & $5.36^{\pm 0.04}$ & $16.54^{\pm 0.18}$ & $16.38^{\pm 2.81}$ \\

& \textbf{Natural PC} & $\mathbf{4.39}^{\pm 0.01}$ & $\mathbf{6.21}^{\pm 0.01}$ & $\mathbf{5.34}^{\pm 0.01}$ & $\mathbf{13.13}^{\pm 0.03}$ & $\mathbf{15.17}^{\pm 0.02}$ \\
\midrule
\multirow{6}{*}{\textbf{Epochs}}
& Standard PC & $15.0^{\pm 0.8}$ & $16.6^{\pm 2.5}$ & $19.4^{\pm 2.1}$ & $2.0^{\pm 0.8}$ & $3.8^{\pm 0.2}$ \\
 & iPC & $12.3^{\pm 0.3}$ & $8.1^{\pm 0.1}$ & $26.8^{\pm 5.9}$ & $\mathbf{0.32}^{\pm 0.0}$ & $2.9^{\pm 0.2}$ \\
  & ePC & $13.4^{\pm 0.4}$ & $32.9^{\pm 12.6}$ & $61.8^{\pm 15.8}$ & $4.4^{\pm 1.0}$ & $7.3^{\pm 5.0}$ \\
 & PC-ALM & $18.9^{\pm 1.2}$ & $24.2^{\pm 1.1}$ & $63.9^{\pm 6.6}$ & $5.3^{\pm 1.3}$ & $4.1^{\pm 0.2}$ \\
 & bPC & $65.0^{\pm 2.5}$ & $46.2^{\pm 1.1}$ & $62.2^{\pm 3.2}$ & $37.5^{\pm 0.3}$ & $24.2^{\pm 10.3}$ \\
& \textbf{Natural PC} & $\mathbf{2.6}^{\pm 0.1}$ & $\mathbf{2.8}^{\pm 0.0}$ & $\mathbf{3.8}^{\pm 0.0}$ & $0.43^{\pm 0.1}$ & $\mathbf{0.29}^{\pm 0.0}$ \\
\midrule
\multirow{6}{*}{\textbf{W-updates}}

& Standard PC & $748^{\pm 40}$ & $831^{\pm 126}$ & $972^{\pm 104}$ & $1{,}024^{\pm 403}$ & $1{,}924^{\pm 117}$ \\
 & iPC & $12.3\text{k}^{\pm 0.3\text{k}}$ & $8{,}060^{\pm 120}$ & $26.8\text{k}^{\pm 5.9\text{k}}$ & $3{,}248^{\pm 396}$ & $28.8\text{k}^{\pm 1.6\text{k}}$ \\
  & ePC & $668^{\pm 20}$ & $1{,}644^{\pm 628}$ & $3{,}088^{\pm 789}$ & $2{,}184^{\pm 502}$ & $3{,}644^{\pm 2{,}484}$ \\
 & PC-ALM & $944^{\pm 62}$ & $1{,}212^{\pm 57}$ & $3{,}196^{\pm 330}$ & $2{,}656^{\pm 661}$ & $2{,}072^{\pm 110}$ \\
 & bPC & $3{,}252^{\pm 124}$ & $2{,}312^{\pm 57}$ & $3{,}112^{\pm 161}$ & $18.7\text{k}^{\pm 0.2\text{k}}$ & $12.1\text{k}^{\pm 5.1\text{k}}$ \\
& \textbf{Natural PC} & $\mathbf{129}^{\pm 6}$ & $\mathbf{139}^{\pm 0}$ & $\mathbf{189}^{\pm 1}$ & $\mathbf{214}^{\pm 27}$ & $\mathbf{143}^{\pm 2}$ \\
\midrule
\multirow{6}{*}{\textbf{TFLOP}}
& Standard PC & $79.7^{\pm 4.3}$ & $88.6^{\pm 13.4}$ & $332^{\pm 35}$ & $2{,}929^{\pm 1{,}153}$ & $2{,}474^{\pm 150}$ \\
& iPC & $\mathbf{2.62}^{\pm 0.07}$ & $\mathbf{1.72}^{\pm 0.03}$ & $\mathbf{18.3}^{\pm 4.0}$ & $\mathbf{18.6}^{\pm 2.3}$ & $\mathbf{74.0}^{\pm 4.1}$ \\
  & ePC& $71.2^{\pm 2.2}$ & $175^{\pm 67}$ & $1{,}053^{\pm 269}$ & $6{,}247^{\pm 1{,}435}$ & $4{,}685^{\pm 3{,}193}$ \\
 & PC-ALM& $101^{\pm 7}$ & $129^{\pm 6}$ & $1{,}090^{\pm 113}$ & $7{,}597^{\pm 1{,}890}$ & $2{,}664^{\pm 141}$ \\
 & bPC & $694^{\pm 26}$ & $493^{\pm 12}$ & $2{,}124^{\pm 110}$ & $107{,}159^{\pm 929}$ & $31{,}062^{\pm 13{,}235}$ \\
 & \textbf{Natural PC} & $23.9^{\pm 1.1}$ & $25.8^{\pm 0.0}$ & $79.3^{\pm 0.4}$ & $629^{\pm 79}$ & $195^{\pm 3}$ \\

\bottomrule
\end{tabular}
\end{adjustbox}
\label{tab:bench-results-gd}
\end{table}

\begin{table}[ht]
\centering
\caption{Performance and cost per method at two widths with regular inference, as Table~\ref{tab:scaling}.}
\small
\begin{adjustbox}{width=\linewidth}
\begin{tabular}{llcccc}
\toprule
\multirow{2}{*}{\textbf{Metric}} & \multirow{2}{*}{\textbf{Method}} & \multicolumn{2}{c}{\textbf{STL-10}} & \multicolumn{2}{c}{\textbf{Tiny ImageNet}} \\
\cmidrule(lr){3-4}\cmidrule(lr){5-6}
& & \textbf{Width 128} & \textbf{Width 256} & \textbf{Width 128} & \textbf{Width 256} \\
\midrule
\multirow{4}{*}{\shortstack[l]{\textbf{MSE}\\($\times 10^{-3}$)}}
 & Standard PC & $13.41^{\pm 0.24}$ & $9.66^{\pm 0.05}$ & $15.26^{\pm 0.03}$ & $11.78^{\pm 0.03}$ \\
 & iPC & $14.90^{\pm 0.04}$ & $9.82^{\pm 0.02}$ & $15.49^{\pm 0.01}$ & $\mathbf{11.72}^{\pm 0.01}$ \\
  & ePC & $13.96^{\pm 0.06}$ & $10.49^{\pm 0.13}$ & $15.79^{\pm 0.48}$ & $12.89^{\pm 0.63}$ \\
 & PC-ALM & $13.35^{\pm 0.11}$ & $9.81^{\pm 0.07}$ & $15.20^{\pm 0.03}$ & $12.13^{\pm 0.04}$ \\
 & bPC & $16.54^{\pm 0.18}$ & $12.01^{\pm 4.64}$ & $16.38^{\pm 2.81}$ & $12.95^{\pm 0.31}$ \\
 & \textbf{Natural PC} & $\mathbf{13.13}^{\pm 0.03}$ & $\mathbf{9.60}^{\pm 0.02}$ & $\mathbf{15.17}^{\pm 0.02}$ & $\mathbf{11.72}^{\pm 0.01}$ \\

\midrule
\multirow{4}{*}{\textbf{Epochs}}
 & Standard PC & $2.05^{\pm 0.81}$ & $2.91^{\pm 0.37}$ & $3.85^{\pm 0.23}$ & $1.67^{\pm 0.17}$ \\
 & iPC & $\mathbf{0.32}^{\pm 0.04}$ & $4.13^{\pm 0.21}$ & $2.88^{\pm 0.16}$ & $1.36^{\pm 0.03}$ \\
  & ePC & $4.37^{\pm 1.00}$ & $5.90^{\pm 0.79}$ & $7.29^{\pm 4.97}$ & $3.53^{\pm 0.64}$ \\
 & PC-ALM & $5.31^{\pm 1.32}$ & $21.26^{\pm 1.42}$ & $4.14^{\pm 0.22}$ & $19.21^{\pm 0.75}$ \\
 & bPC & $37.46^{\pm 0.32}$ & $28.74^{\pm 13.77}$ & $24.15^{\pm 10.29}$ & $36.13^{\pm 1.15}$ \\
 & \textbf{Natural PC} & $0.43^{\pm 0.05}$ & $\mathbf{0.46}^{\pm 0.02}$ & $\mathbf{0.29}^{\pm 0.00}$ & $\mathbf{0.42}^{\pm 0.02}$ \\

\midrule
\multirow{4}{*}{\textbf{W-updates}}
 & Standard PC & $1{,}024^{\pm 403}$ & $1{,}456^{\pm 184}$ & $1{,}924^{\pm 117}$ & $836^{\pm 87}$ \\
 & iPC & $3{,}248^{\pm 396}$ & $41.28\text{k}^{\pm 2.06\text{k}}$ & $28.80\text{k}^{\pm 1.60\text{k}}$ & $13.60\text{k}^{\pm 0.25\text{k}}$ \\
  & ePC & $2{,}184^{\pm 502}$ & $2{,}952^{\pm 393}$ & $3{,}644^{\pm 2{,}484}$ & $1{,}764^{\pm 319}$ \\
 & PC-ALM & $2{,}656^{\pm 661}$ & $10.6\text{k}^{\pm 0.7\text{k}}$ & $2{,}072^{\pm 110}$ & $9{,}604^{\pm 373}$ \\
 & bPC & $18.7\text{k}^{\pm 0.2\text{k}}$ & $14.4\text{k}^{\pm 6.9\text{k}}$ & $12.1\text{k}^{\pm 5.1\text{k}}$ & $18.1\text{k}^{\pm 0.6\text{k}}$ \\
 & \textbf{Natural PC} & $\mathbf{214}^{\pm 27}$ & $\mathbf{228}^{\pm 8}$ & $\mathbf{143}^{\pm 2}$ & $\mathbf{209}^{\pm 9}$ \\

\midrule
\multirow{4}{*}{\shortstack[l]{\textbf{Updates to}\\\textbf{Std. PC plateau}}}
 & Standard PC & $780^{\pm 51}$ & $1{,}468^{\pm 131}$ & $1{,}924^{\pm 79}$ & $820^{\pm 22}$ \\
 & iPC & never & $82\text{k}^{\pm 10.13\text{k}}$ & $65\text{k}^{\pm 1.94\text{k}}$ & $12\text{k}^{\pm 0.49\text{k}}$ \\
  & ePC & never & never & $7{,}670^{\pm 550}$\, & never \\
 & PC-ALM & $2{,}164^{\pm 259}$ & $12.4\text{k}^{\pm 0.6\text{k}}$\, & $1{,}924^{\pm 48}$ & never \\
 & bPC & never & $16.9\text{k}^{\pm 2.1\text{k}}$\, & $12.9\text{k}^{\pm 0.8\text{k}}$\, & never \\
 & \textbf{Natural PC} & $\mathbf{133}^{\pm 8}$ & $\mathbf{192}^{\pm 8}$ & $\mathbf{130}^{\pm 4}$ & $\mathbf{186}^{\pm 4}$ \\

\midrule
 \multirow{4}{*}{\textbf{TFLOP}}
  & Standard PC & $2{,}929^{\pm 1{,}153}$ & $8{,}367^{\pm 1{,}057}$ & $2{,}474^{\pm 150}$ & $2{,}172^{\pm 226}$ \\
   & iPC & $\mathbf{18.6}^{\pm 2.3}$ & $\mathbf{474}^{\pm 24}$ & $\mathbf{74.0}^{\pm 4.1}$ & $\mathbf{70.6}^{\pm 1.3}$ \\
    & ePC & $6{,}247^{\pm 1{,}435}$ & $16{,}964^{\pm 2{,}259}$ & $4{,}685^{\pm 3{,}193}$ & $4{,}583^{\pm 830}$ \\
 & PC-ALM & $7{,}597^{\pm 1{,}890}$ & $61{,}076^{\pm 4{,}072}$ & $2{,}664^{\pm 141}$ & $24{,}949^{\pm 970}$ \\
 & bPC & $107{,}159^{\pm 929}$ & $165{,}224^{\pm 79{,}144}$ & $31{,}062^{\pm 13{,}235}$ & $93{,}878^{\pm 2{,}985}$ \\
& \textbf{Natural PC} & $629^{\pm 79}$ & $1{,}364^{\pm 48}$ & $195^{\pm 3}$ & $592^{\pm 26}$ \\

\bottomrule
\end{tabular}
\end{adjustbox}
\label{tab:scaling-gd}
\end{table}

\newpage
\section{Image Classification}
\label{app:cls_results}

We evaluate all methods on supervised classification with the standard discriminative construction of predictive coding \citep{whittington2017approximation,pinchetti2025benchmarkingpredictivecodingnetworks}. The image is clamped at the deepest layer and the one-hot label is the observation. During learning, the hidden states are initialized by a forward sweep and relax for $50$ fixed-step iterations with both ends clamped. At test time the prediction is the forward sweep, which coincides with the energy minimum when the output is free. Since the contribution of this work is the geometry of the weight update, every other choice is kept fixed and minimal.

Results are reported in Table ~\ref{tab:cls_full}. All PC methods update the weights with AdamW, learning rate schedules or output nudging, and use no weight decay, no biases, tanh activations and a learning rate matched per dataset. Standard PC is the plain Hebbian rule of \eqref{eq:pc-update}. The BP baseline uses AdamW on the same architecture and squared-error loss. The L-BFGS baseline uses the two-loop recursion with a history of $10$ pairs for minibatch and $20$ for full-batch training, curvature pairs from the same batch, the cautious acceptance rule $s^\top y > \epsilon,|s|^2$, the initial scaling $H_0 = (s^\top y / y^\top y),I$, an Armijo line search starting from unit step length, a steepest-descent fallback when no pair is accepted, and no weight decay. Every run receives the same budget of $250$ epochs without early stopping, because adaptive stopping criteria assign different budgets to different optimizers and can produce apparent stalls or collapses. Accuracies are reported at the end of the shared budget. Networks are three-layer perceptrons of width $256$ trained on the full training sets.

Additionally, we report depth scaling results on the exact same experimental setup in Table ~\ref{tab:cls-depth-scaling}. Diverged indicates that at least 2 of 5 seeds diverged during training. Mean $\pm$ std are reported over 5 seeds. The learning rate was individually tuned.

\begin{table}[!htbp]
\centering
\caption{Classification accuracy at a fixed budget of $250$ epochs (width $256$, depth $3$, tanh. Mean $\pm$ standard deviation over three seeds). Accuracy is the test accuracy at the end of the budget. The speed columns give the epochs and weight updates needed to first reach a reference accuracy of $90$, $80$ and $30$ percent on MNIST, FashionMNIST and CIFAR-10 respectively. $^a$Full-batch L-BFGS runs $250$ full-data iterations, about $2.7$ times the compute of the other methods. At matched compute of about $90$ iterations, the accuracies are $86.7$, $76.4$ and $15.4$ percent.}
\vspace{0.6em}
\small
\begin{tabular}{lllll}
\toprule
\multirow{2}{*}{\textbf{Task}} & \multirow{2}{*}{\textbf{Method}} & \textbf{Acc.\ (\%)} &
\textbf{Epochs} & \textbf{W-updates} \\ & & \textbf{@250 ep} & \textbf{$\to$ ref.} & \textbf{$\to$ ref.} \\ \midrule
\multirow{6}{*}{MNIST}
 & Standard PC             & $94.71^{\pm 0.26}$                 & $31.9^{\pm 4.2}$             & $10\text{k}^{\pm 1\text{k}}$ \\
 & \textbf{Natural PC}     & $\mathbf{97.72}^{\pm 0.12}$         & $\mathbf{1.6}^{\pm 0.2}$     & $473^{\pm 45}$ \\
 & iPC                     & $85.01^{\pm 0.10}$                 & never                        & never \\
 & \textbf{Backprop (AdamW)} & $97.24^{\pm 0.12}$               & $\mathbf{1.6}^{\pm 0.2}$     & $\mathbf{467}^{\pm 47}$ \\
 & Backprop (L-BFGS)       & $93.98^{\pm 0.18}$                 & $10.7^{\pm 1.9}$             & $3\text{k}^{\pm 1\text{k}}$ \\
 & Backprop (L-BFGS, full) & $93.36^{\pm 0.16}$\,$^{a}$         & $105.3^{\pm 4.0}$\,$^{b}$    & $105^{\pm 4}$ \\

\midrule
\multirow{6}{*}{FMNIST}
 & Standard PC             & $70.05^{\pm 17.08}$                & $26.0^{\pm 6.6}$             & $8\text{k}^{\pm 2\text{k}}$ \\
 & Natural PC              & $88.62^{\pm 0.13}$                 & $2.2^{\pm 0.5}$              & $680^{\pm 115}$ \\
 & iPC                     & $80.60^{\pm 0.09}$                 & $105.9^{\pm 3.5}$            & $635\text{k}^{\pm 21\text{k}}$ \\
 & \textbf{Backprop (AdamW)} & $\mathbf{88.63}^{\pm 0.03}$       & $\mathbf{1.7}^{\pm 0.3}$     & $500^{\pm 82}$ \\
 & Backprop (L-BFGS)       & $83.25^{\pm 0.05}$                 & $4.5^{\pm 0.4}$              & $1\text{k}$ \\
 & Backprop (L-BFGS, full) & $81.81^{\pm 0.02}$\,$^{a}$         & $59.7^{\pm 3.1}$\,$^{b}$     & $\mathbf{60}^{\pm 3}$ \\

\midrule
\multirow{6}{*}{CIFAR}
 & Standard PC             & $31.28^{\pm 1.32}$                 & $35.5^{\pm 7.3}$             & $9\text{k}^{\pm 2\text{k}}$ \\
 & \textbf{Natural PC}     & $\mathbf{48.03}^{\pm 0.45}$         & $\mathbf{2.4}^{\pm 0.1}$     & $\mathbf{353}^{\pm 21}$ \\
 & iPC                     & $29.11^{\pm 0.72}$                 & never                        & never \\
 & Backprop (AdamW)        & $45.15^{\pm 0.43}$                 & $6.5^{\pm 0.2}$              & $2\text{k}$ \\
 & Backprop (L-BFGS)       & $30.21^{\pm 0.94}$                 & $14.4^{\pm 2.1}$             & $4\text{k}^{\pm 1\text{k}}$ \\
 & Backprop (L-BFGS, full) & $21.60^{\pm 7.71}$\,$^{a}$         & never                        & never \\
\bottomrule
\end{tabular}
\label{tab:cls_full}
\end{table}

\begin{table}[t]
\centering
\caption{Classification Accuracy by Depth, Activation Function and Method}
\vspace{0.6em}
\small
\setlength{\tabcolsep}{6pt}
\begin{tabular}{l cccc}
\toprule
Method & $d{=}4$ & $d{=}8$ & $d{=}32$ & $d{=}64$ \\
\midrule
\multicolumn{5}{c}{\textbf{tanh}} \\
\midrule
Standard PC & $87.1^{\pm1.1}$ & $82.1^{\pm0.5}$ & $34.7^{\pm0.8}$ & $25.2^{\pm1.1}$ \\
iPC & $33.3^{\pm0.5}$ & \emph{diverged} & $10.9^{\pm0.4}$ & $10.7^{\pm0.1}$ \\
Implicit Natural PC & $70.1^{\pm0.8}$ & \emph{diverged} & \emph{diverged} & \emph{diverged} \\
Natural PC (fixed lr) & $94.5^{\pm0.2}$ & $91.1^{\pm0.8}$ & $36.6^{\pm0.7}$ & $22.6^{\pm2.2}$ \\
Natural PC (Lipschitz step) & $97.1^{\pm0.1}$ & $96.7^{\pm0.2}$ & $\mathbf{77.7}^{\pm9.5}$ & $\mathbf{50.8}^{\pm1.3}$ \\
Backprop (AdamW) & $\mathbf{97.7}^{\pm0.1}$ & $\mathbf{97.5}^{\pm0.1}$ & \emph{diverged} & \emph{diverged} \\
Backprop (L-BFGS, full batch) & $89.3^{\pm2.7}$ & $90.4^{\pm0.3}$ & $77.2^{\pm6.8}$ & \emph{diverged}\\
\midrule
\multicolumn{5}{c}{\textbf{ReLU}} \\
\midrule
Standard PC & $87.7^{\pm0.2}$ & $67.3^{\pm5.5}$ & $15.5^{\pm1.1}$ & $12.7^{\pm1.2}$ \\
iPC & $19.0^{\pm0.3}$ & \emph{diverged} & $9.9^{\pm0.4}$ & $10.2^{\pm0.2}$ \\
Implicit Natural PC & $20.2^{\pm2.7}$ & $11.1^{\pm0.2}$ & $10.6^{\pm0.3}$ & $10.1^{\pm0.2}$ \\
Natural PC (fixed lr) & $89.6^{\pm0.4}$ & $72.8^{\pm1.2}$ & $18.8^{\pm1.5}$ & $14.6^{\pm1.3}$ \\
Natural PC (Lipschitz step) & $93.1^{\pm0.4}$ & $81.0^{\pm1.1}$ & $23.3^{\pm0.9}$ & $\mathbf{15.9}^{\pm1.8}$ \\
Backprop (AdamW) & $\mathbf{97.2}^{\pm0.1}$ & $\mathbf{97.2}^{\pm0.1}$ & $\mathbf{94.9}^{\pm1.6}$ & \emph{diverged} \\
Backprop (L-BFGS, full batch) & $92.9^{\pm4.4}$ & $96.8^{\pm0.5}$ & \emph{diverged} & \emph{diverged} \\
\bottomrule
\end{tabular}
\label{tab:cls-depth-scaling}
\end{table}

\newpage
\section{Implicit Natural PC}
\label{app:implicit-natural-pc}
Natural PC inverts the presynaptic correlations that defines the natural gradient and replaces the equilibrium activations that carry the correlations with this filtered learning signal. However, decorrelating the activity directly during inference offers a second route to natural gradient descent. When the equilibrium Gram $G_l^*$ is diagonal, the preconditioner in \eqref{eq:nat-grad-target} reduces to a per-coordinate rescaling, and the standard PC update of \eqref{eq:pc-update} already points along the natural gradient up to a scale. A lateral circuit that whitens the activity during inference should therefore make natural gradient learning implicit, without changing the weight update rule. This variant follows the view that lateral inhibition decorrelates the representation \citep{king2013inhibitory,chettih2019single,znamenskiy2024functional}, and we refer to it as Implicit Natural PC.

Following the recurrent whitening network of \citet{pehlevan2015optimization}, we add to each layer a symmetric lateral weight matrix $B_l$ that contributes $\tfrac{1}{2}\phi(z_l)^{\top}B_l\phi(z_l)$ to the energy of \eqref{eq:freeenergy}, so that the activities settle under lateral competition. The lateral weights follow the anti-Hebbian rule
$$B_l \leftarrow \mathrm{clip}\bigl(B_l + \eta_B,(G_l - c,I)\bigr), \qquad c = \mathrm{mean},\mathrm{diag},G_l,$$
where $G_l$ is the Gram of the equilibrium activity on the current batch. The rule increases competition between co-active units and decreases competition between anti-correlated units, and reaches a fixed point when $G_l = cI$, that is, when the activity is white up to a scale. Because $B_l$ enters the Hessian of the inference dynamics, we clip the eigenvalues of $B_l$ to a bounded interval so that the fixed-step relaxation remains stable across layer widths. The weight update remains the standard PC rule of \eqref{eq:pc-update}.

\begin{figure}[ht]
\centering
\includegraphics[width=0.9\linewidth]{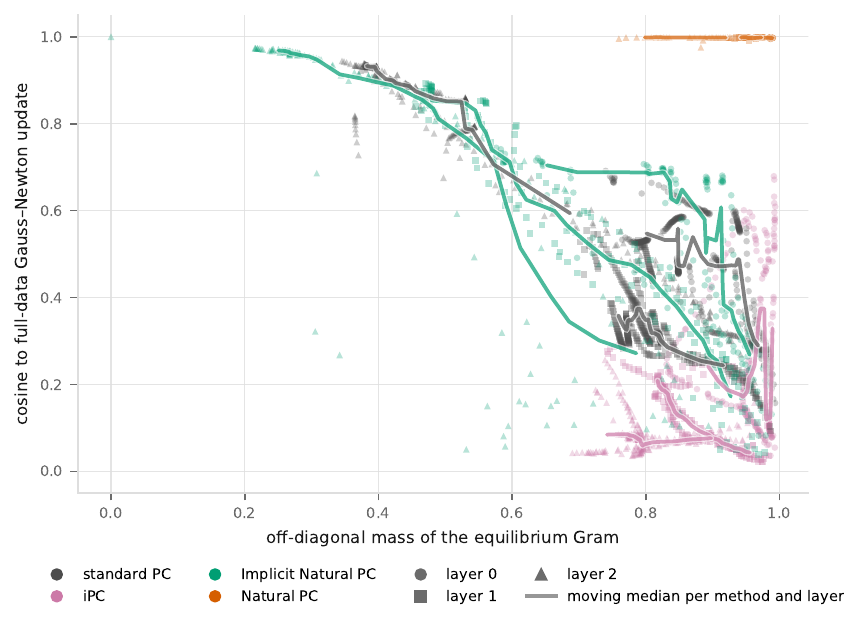}
\caption{Alignment of each method's weight update with the natural gradient, plotted against the off-diagonal mass of the equilibrium Gram. Points are training checkpoints pooled over MNIST, FashionMNIST and CIFAR-10. Marker shape encodes the layer, colour the method, and heavy curves are moving medians per method and layer. The reference update uses the minibatch gradient with the full-data Gram.}
\label{fig:gram-structure}
\end{figure}

In Fig.~\ref{fig:gram-structure}, we plot the alignment of the applied update with the natural gradient against the off-diagonal mass of the equilibrium Gram, pooled over training checkpoints, layers and datasets. Standard PC aligns with the natural gradient only while the Gram is close to diagonal and falls to a cosine of about $0.2$ as correlations increase. Implicit Natural PC lies above standard PC at matched off-diagonal mass, so the whitening improves alignment beyond what a more diagonal Gram alone would give. Natural PC holds a cosine above $0.99$ at off-diagonal masses of $0.8$ to $1.0$, where every method without an explicit inverse has fallen to $0.2$ to $0.5$. Diagonality is a requirement only for updates that do not invert the Gram.

\newpage
\section{Analysis of the Lateral Circuit}
\label{app:dissecting-circ}

The main text shows that Natural PC learns faster and reaches lower error than the other PC methods. This appendix asks what the lateral circuit computes and why the computation helps. We examine what the lateral weights store, how the learned metric evolves, how much temporal averaging the circuit needs and whether the speedup is caused by the circuit online. Throughout, Natural PC denotes the lateral-circuit model of the main text, and Implicit Natural PC the whitened-inference variant of App.~\ref{app:implicit-natural-pc}. All experiments use the width-64 reconstruction setting with shared initialization and batch order per seed. Curve panels report three seeds, with bands showing the range across seeds $\pm$ std, and landscape, heatmap and eigenmode panels show one representative seed. Misalignment is measured as $1-\cos(\cdot,,U(g,G_N^*))$, where $U(g,G) = g,(G+\lambda I)^{-1}$ is the natural gradient update with the full-data Gram, evaluated at each run's own weights on a fixed probe set.

\subsection{What do the lateral weights store? }
\begin{figure}[ht]
\centering
\includegraphics[width=\linewidth]{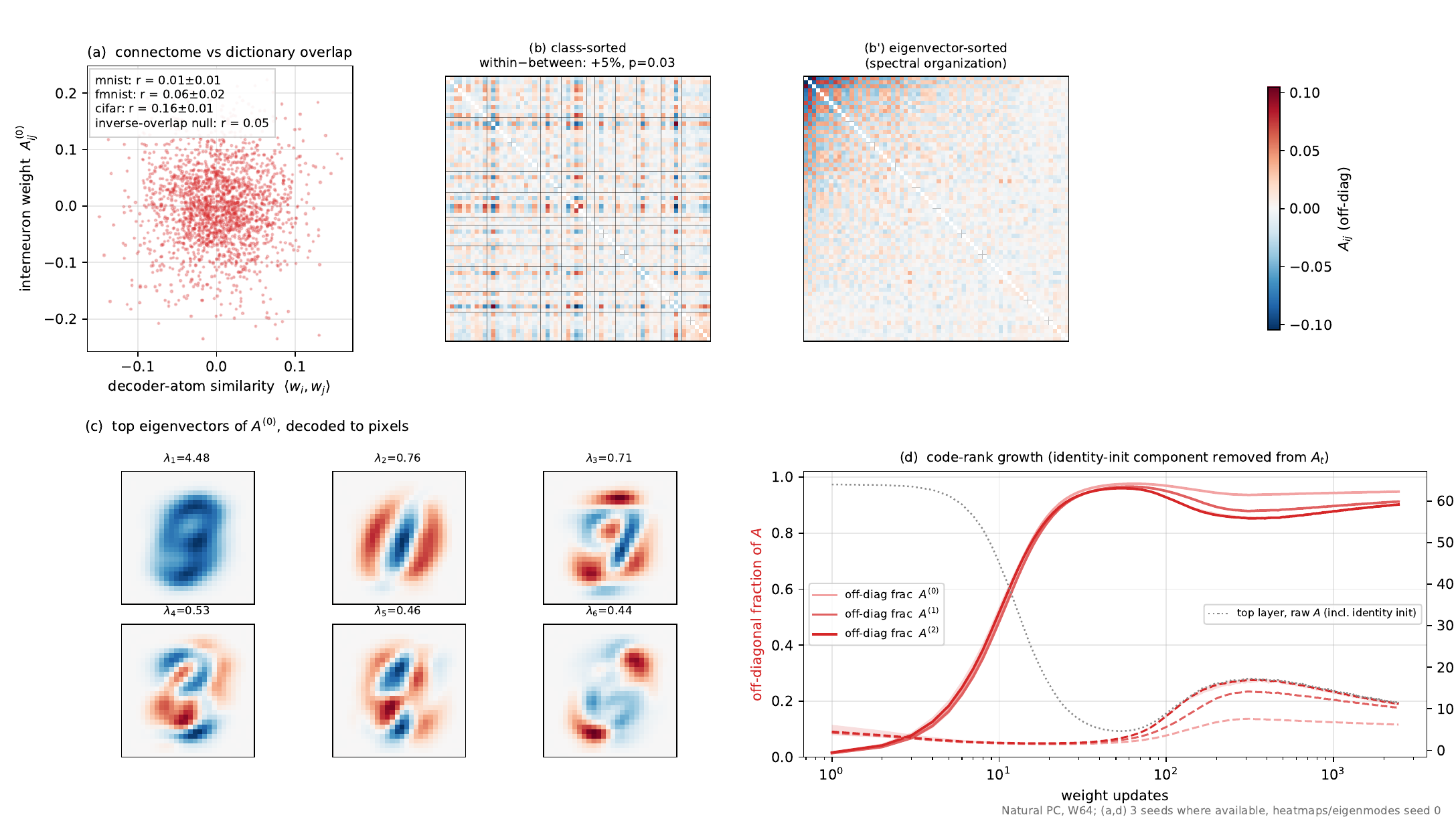}
\caption{Structure of the learned lateral weights (Natural PC, width 64, MNIST unless stated). (a) Lateral weight $A^{(0)}_{ij}$ against the similarity of the corresponding decoder columns, with correlations for three datasets and the inverse-overlap prediction. (b) Final top-layer $A$ sorted by class, with the within-class excess from a permutation test. (b') The same matrix sorted by loading on the leading eigenvector. (c) The six leading eigenvectors of $A^{(0)}$ decoded to pixel space, with eigenvalues. (d) Off-diagonal fraction of $A$ over training for the three layers, after removing the identity initialization (solid, left axis), and participation ratio (dashed, right axis); the dotted line shows the raw top-layer matrix including the initialization.}
\label{fig:app-connectome}
\end{figure}

The lateral weights of the first hidden layer, $A^{(0)}$, connect interneurons that are driven by principal neurons with decoder columns $w_i$ and $w_j$. A hypothesis is that the weight between two interneurons follows how similar or how overlapping the two decoder columns are. We falsify these hypotheses (Fig.~\ref{fig:app-connectome}a). The correlation between $A^{(0)}_{ij}$ and the decoder similarity $\langle w_i, w_j\rangle$ is $0.01$ on MNIST, $0.06$ on FashionMNIST and $0.16$ on CIFAR-10, and the correlation with the inverse-overlap prediction is $0.05$. Moreover, digit classes also do not predict the structure of the lateral weights. When the matrix is sorted by digit class, weights between interneurons of the same class exceed weights between classes by only $5$ percent (Fig.~\ref{fig:app-connectome}b).

Sorting by the leading eigenvector instead exposes the structure that is present (Fig.~\ref{fig:app-connectome}b'). Most of the variance in the lateral weights is account for by a few eigenvectors that span the whole population. The leading eigenvalue is six times the second, and the leading eigenvectors decode to image-scale patterns, a mean-digit blob followed by stroke-like components (Fig.~\ref{fig:app-connectome}c). These eigenvectors form within the first tens of updates. After removing the identity initialization, the off-diagonal fraction of $A$ rises from near zero to about $0.9$ within the first $50$ updates in all three layers, while the effective rank, measured as the participation ratio of the eigenvalues, grows from about $2$ to about $17$ over training (Figure~\ref{fig:app-connectome}d). The lateral weights therefore store the main directions along which the principal neurons vary together, which is a statistical property of the activity and is unrelated to the features the principal neurons encode.

\subsection{How is the Gram matrix learned?}
\begin{figure}[ht]
\centering
\includegraphics[width=\linewidth]{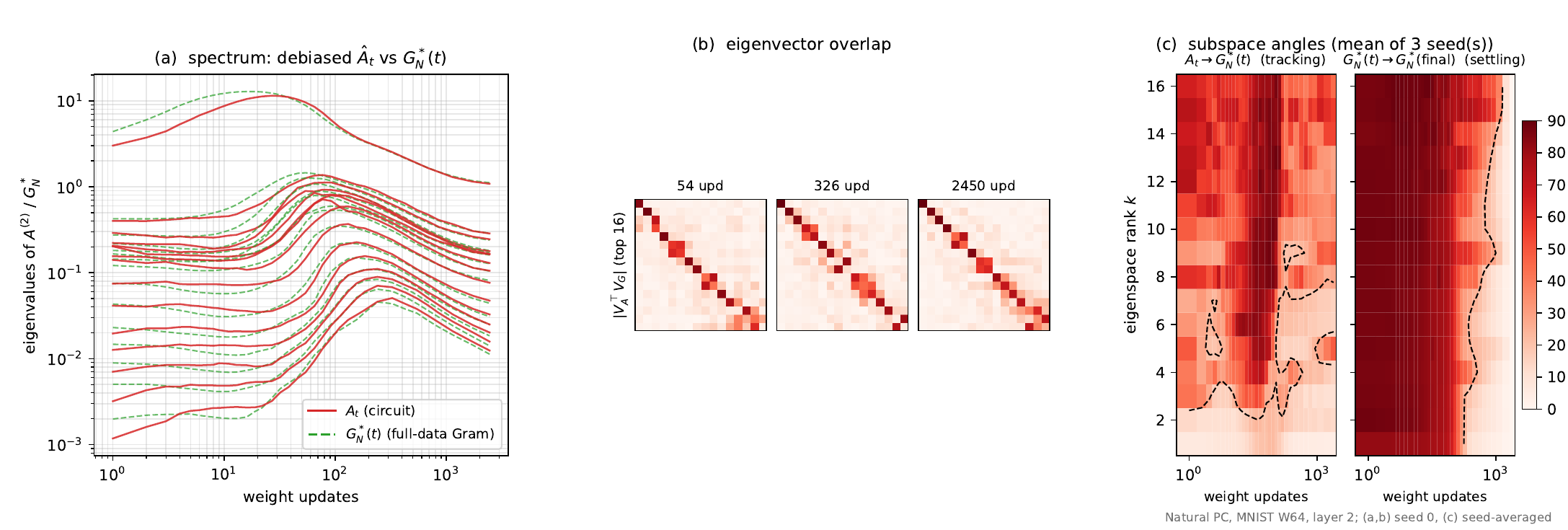}
\caption{Learning of the equilibrium Gram (Natural PC, MNIST, width 64, top layer). (a) Eigenvalues of the debiased lateral matrix $\hat A_t$ (red) and of the full-data equilibrium Gram $G_N^*(t)$ (green, dashed) over training. (b) Absolute overlap between the top sixteen eigenvectors of the two matrices at three times. (c) Principal angles between the top-$k$ eigenspaces, averaged over three seeds. Left: lateral matrix against the Gram at the same time. Right: Gram at time $t$ against the final Gram. Dashed contours mark $45$ degrees.}
\label{fig:app-spectrum}
\end{figure}

The lateral weights are an estimate of a target that moves during training, since the equilibrium Gram $G_N^*(t)$ changes as the weights change. Fig.~\ref{fig:app-spectrum} compares the debiased lateral matrix $\hat A_t$, from which the decayed identity initialization $(1-\alpha)^t I$ has been removed, of the top layer with the full-data Gram at the same time. The eigenvalues of the two matrices agree closely at every rank and at every time (Fig.~\ref{fig:app-spectrum}a), even though the spectrum itself changes by more than an order of magnitude: the leading eigenvalue rises during the first $30$ updates and then decays, and the eigenvalues of ranks two to sixteen rise sharply between $50$ and $300$ updates as the code develops structure. The eigenvectors of the two matrices are aligned as well, with the overlap matrix of the top sixteen eigenvectors close to diagonal at $54$, $326$ and $2{,}450$ updates (Fig.~\ref{fig:app-spectrum}b). The subspace angles in Fig.~\ref{fig:app-spectrum}c separate two processes. The left panel shows how closely the lateral matrix at time $t$ matches the Gram at that same time point. The right panel shows how the metric settles toward the final Gram: the leading eigenvector stops rotating after about $500$ updates and the full sixteen-dimensional subspace after about $10^3$. In between, the angles for ranks three and higher are large, which is the phase in which many eigenvalues cross (Fig.~\ref{fig:app-spectrum}a), so individual eigenvectors are poorly defined and the top-$k$ subspaces of two nearly identical spectra can differ. The lateral weights therefore follow a moving metric rather than converging to a fixed one, and the eigenvalues are tracked throughout, while eigenvector tracking is limited only where the metric itself is degenerate.

\subsection{The speedup is caused by the lateral circuit acting online}
\begin{figure}[ht]
\centering
\includegraphics[width=\linewidth]{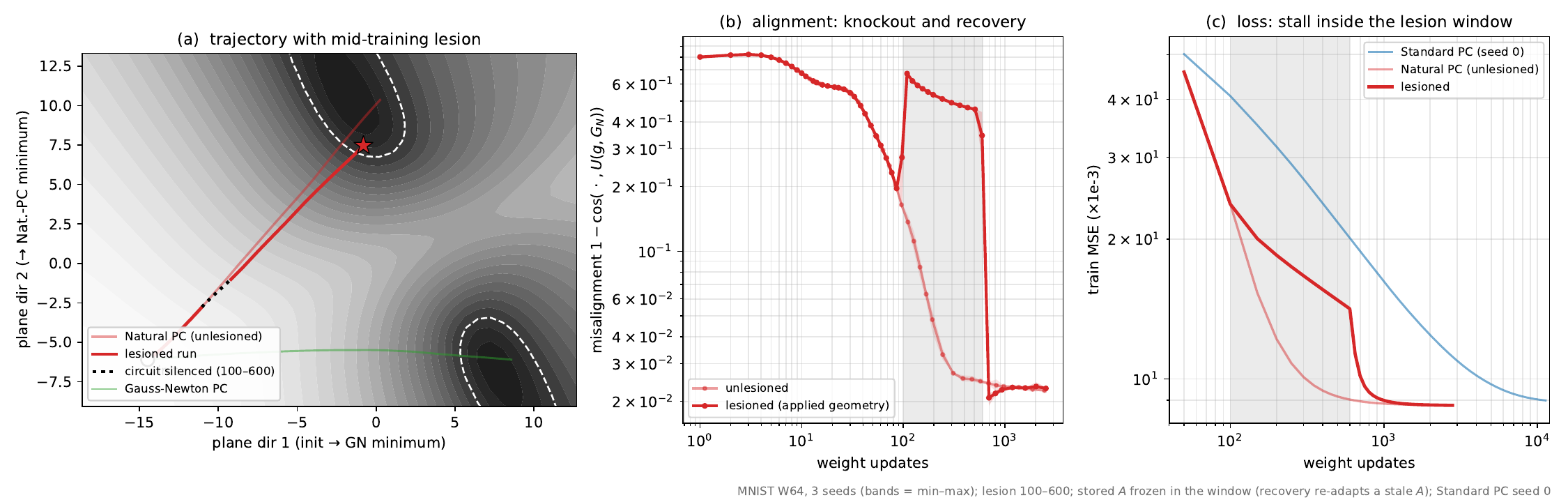}
\caption{Mid-training lesion of the interneurons (Natural PC, MNIST, width 64, three seeds; bands show the range across seeds). The interneurons are silenced between updates $100$ and $600$ (shaded), with the lateral weights frozen. (a) Trajectories over the equilibrium loss in the plane spanned by the initialization, the Gauss-Newton PC minimum and the Natural PC minimum; the silenced segment is dotted. (b) Misalignment of the applied update with the full-data natural gradient for the lesioned and unlesioned runs. (c) Training error of the lesioned run compared with unlesioned Natural PC and with standard PC.}
\label{fig:app-lesion}
\end{figure}

Does Natural PC converge faster because the interneurons shape every weight update, or because their early influence puts the weights on a better path? To separate the two, we silence the interneurons between updates $100$ and $600$ while keeping the lateral weights frozen at their values from update $100$, so that the weight update falls back to the standard PC rule (Fig.~\ref{fig:app-lesion}). The applied update changes geometry instantly. Misalignment with the natural gradient jumps from about $0.2$ to about $0.65$ at the onset of the lesion and decays only slowly within the window (Fig.~\ref{fig:app-lesion}b). The loss responds in the same way. Inside the lesion window, the training error decreases at the rate of standard PC, which is visibly slower than the unlesioned run (Fig.~\ref{fig:app-lesion}c). When the interneurons are restored at update $600$, the update realigns with the natural gradient within a few tens of updates, although the lateral weights are $500$ updates out of date, and the loss drops to the unlesioned trajectory. In weight space the lesioned run leaves the unlesioned path during the window and returns to the same minimum afterwards (Fig.~\ref{fig:app-lesion}a). The acceleration is therefore produced by the interneurons at the time of each update, and the stored correlations remain useful over hundreds of updates.

 \end{document}